\documentclass{article}

\usepackage[preprint]{neurips_2026}

\usepackage[utf8]{inputenc}
\usepackage[T1]{fontenc}
\usepackage{hyperref}
\usepackage{url}
\usepackage{booktabs}
\usepackage{amsfonts}
\usepackage{amsmath}
\usepackage{amssymb}
\usepackage{nicefrac}
\usepackage{microtype}
\usepackage{xcolor}
\usepackage{graphicx}
\usepackage{tikz}
\usetikzlibrary{positioning,arrows.meta}
\usepackage{float}

\renewenvironment{abstract}{%
  \vskip 0.075in%
  \centerline{\large\bf Abstract}%
  \vspace{0.5ex}%
}{%
  \par%
  \vskip 1ex%
}

\title{When Tabular Foundation Models Transfer Across Modalities: A Systematic Evaluation Across 95 Datasets, 7 Modalities, and Two Regimes}

\author{%
  Julien Lafrance\\
  T\'el\'ecom Paris, Institut Polytechnique de Paris\\
  Palaiseau, France\\
  \texttt{julien.lafrance@telecom-paris.fr}
}

\begin{document}

\maketitle

\begin{abstract}
We present a single classification pipeline that combines an Equiangular Tight Frame (ETF) preprocessing stage with a tabular foundation model for in-context inference, applied identically across modalities once data is mapped to fixed vector representations. We evaluate it on 95 datasets spanning seven signal modalities -- vision, audio, speech, text, molecular, time-series, and tabular. The main methodological contribution is to fix the comparison object: throughout the paper, performance is judged against the strongest lightweight tuned baseline on the same frozen features, while oracle selection, deployed selection, and specialized fine-tuning are reported separately.

The pipeline is broadly competitive with strong lightweight tuned baselines on the same frozen features. It does not match the very best specialized models or heavily tuned pipelines on every task, but it stays close, and it runs much faster -- typically 4 to 200 times faster than full backbone fine-tuning, often at comparable quality.

We describe how to deploy the pipeline in practice: when to apply ETF preprocessing, how to stop its training without a validation split, how to set up the in-context classifier, and how to calibrate the resulting probabilities. The calibration step is non-cosmetic: TabICL produces well-calibrated probabilities by construction, ETF preprocessing initially disrupts that calibration, and the post-hoc rescaling restores it -- yielding a per-prediction confidence signal that practitioners can use as a trust threshold for confidence-gated deployment. We also report where the pipeline should not be expected to help, and how to identify those cases in advance.
\end{abstract}

\section{Introduction}\label{sec:introduction}

Classification is fragmented by modality. Tabular data goes to gradient-boosted trees. Vision and text embeddings go to tuned linear probes. Molecules go to ChemBERTa fine-tuning, audio to AST, graphs to end-to-end GIN training. Each modality has its own tooling, its own conventions, its own tuning recipes \citep{Chen2016, Kornblith2019, Chithrananda2020, Gong2021, Xu2018}.

This is costly to maintain, and it makes the literature on "cross-modality transfer" hard to read. A claim that a method "transfers" can mean almost anything: a fixed-feature comparison, a full fine-tuning comparison, an oracle selection over many variants. Critical reviews of graph benchmarks have shown how easily gains dissolve under stricter protocols \citep{Errica2020, Tonshoff2023}.

We take a simple position: once data is mapped to a fixed vector representation, the downstream classification step does not need to be modality-specific. We test that position with one pipeline applied identically to 95 datasets across seven modalities, under a deliberately strict protocol -- the comparison is always against the strongest lightweight tuned baseline on the same frozen features.

The pipeline has three stages: an adaptive Equiangular Tight Frame (ETF) preprocessing step that reshapes the feature geometry, in-context classification with TabICL \citep{Qu2025, Qu2026}, and post-hoc temperature scaling. An equiangular tight frame is a set of vectors that are maximally and equally separated -- all pairwise angles equal and as large as the dimension allows -- which is the geometry that class means approach during neural collapse \citep{Papyan2020}; our preprocessing step encourages the feature representation toward this geometry before classification. Tabular foundation models like TabPFN \citep{Hollmann2025} and TabICL classify vector inputs through pretrained in-context inference, which makes them natural candidates for a common downstream engine. We use TabICL rather than TabPFN because it converges to comparable accuracy on our panels at roughly $3\times$ lower inference cost, with a more permissive open license (Apache~2.0); the choice does not affect the protocol or the headline conclusions. Prior work has tested both on individual modalities \citep{Pinto2025, BenHicham2026}, but not as a uniform pipeline across many.

What we report:

\begin{itemize}
\item How the pipeline performs against strong lightweight tuned baselines on the same frozen features, and against fully fine-tuned specialized models.
\item Two clear use cases that emerge from the data: tasks where the pipeline matches the baseline and saves a lot of time, and tasks where it actually beats the baseline. We also give a simple pre-deployment rule to distinguish them in advance.
\item How to use the pipeline in practice: the four key decisions (preprocessing, training stop, ensembling, calibration) and where each comes from.
\item Where the pipeline does not help, and how to recognize those cases.
\end{itemize}

We position the work against the tabular foundation model, frozen-feature transfer, and neural-collapse literatures in Section~\ref{sec:related} (with extended discussion in Appendix~\ref{app:related_extended}). Our specific contributions there: the first systematic evaluation of a tabular foundation model on a wide diversity of frozen encoder embeddings (10+ encoder families across 7 modalities), ETF preprocessing as a modular pre-collapse stage rather than as a property of encoder pre-training, its first combination with a tabular foundation model classifier, and a validation-free geometric early-stopping criterion that we audit on eight controlled trajectories.

\section{Related work and positioning}\label{sec:related}

\textbf{Tabular foundation models on non-tabular data.} TabPFN \citep{Hollmann2025} introduced the in-context-inference paradigm for tabular classification; TabICL \citep{Qu2025, Qu2026} scaled it with architectural improvements and a more diverse synthetic prior. A growing $2025$--$2026$ literature has begun adapting these models beyond classical tabular data: \citet{Pinto2025} and \citet{BenHicham2026} on molecular embeddings, \citet{Hayler2025} on graph node classification via flattening, and the contemporaneous MultiModalPFN \citep{Kim2026} (CVPR~2026) on paired tabular-plus-image or tabular-plus-text inputs with light fine-tuning on $9$ datasets. Each restricts itself to a single non-tabular modality or a paired bi-modal setting; to our knowledge no work systematically evaluates a tabular FM on a wide diversity of frozen, single-modality encoder embeddings under a unified zero-touch protocol. That cross-modality stress test, on $12$ encoder families across $7$ modalities, is the target of this paper.

\textbf{Methodological context.} Our pipeline combines two ingredients that have not, to our knowledge, been combined before: ETF preprocessing as a modular pre-collapse stage, and a tabular foundation model as a zero-touch downstream classifier. The protocol is a same-features comparison in the strict sense of \citet{Kornblith2019}: we keep the frozen features and replace only the classification head, so any improvement is attributable to the head. The methodological lesson that cross-modality claims require fixing the comparison object before they are meaningful transfers from graph learning \citep{Errica2020}. The ETF preprocessing borrows from neural collapse \citep{Papyan2020, Han2024, Yang2022}, whose connection to transfer was studied by \citet{Galanti2022} and \citet{Li2023NC} (``preventing within-class collapse to a certain extent leads to better transferability''), both keeping collapse modulation \emph{internal} to encoder pre-training. We use it as a \emph{modular} preprocessing stage instead, encoder-agnostic, with a geometric early-stopping criterion (the within/between-class scatter ratio $R$) operating purely on training-side statistics. Post-hoc temperature scaling \citep{Guo2017} is used off-the-shelf; the empirical observation that TabICL produces probabilities well-calibrated by construction, that ETF preprocessing disrupts that calibration, and that rescaling restores it well enough to support confidence-gated deployment is, to our knowledge, the first such characterization in the cross-modality classification setting.

\section{The pipeline and how we evaluate it}\label{sec:the_pipeline_and_how_we_evaluate_it}

\subsection{The pipeline}\label{sub:the_pipeline}

The pipeline has three stages, applied in order:

\begin{enumerate}
\item \textbf{ETF preprocessing.} A small encoder maps the input features to a 256-dimensional space where classes are encouraged toward an equiangular tight frame geometry. Skipped when the input dimension is already small ($d \le 30$), because at that scale the preprocessing overhead outweighs the geometric gain.
\item \textbf{TabICL inference.} A pretrained tabular foundation model classifies new points by in-context comparison with the training set. We use ensembling over 8 context permutations on raw features and a single permutation on ETF-preprocessed features (the ensembling gain mostly disappears after ETF, see Appendix~\ref{app:mechanism}).
\item \textbf{Temperature scaling.} ETF outputs are overconfident. We rescale the logits using a temperature fitted on a 20\% holdout of the training set \citep{Guo2017}.
\end{enumerate}

\textbf{ETF training objective.} The ETF preprocessing stage trains a depth-4 MLP $\theta$ that maps encoder features $x$ to a $256$-dimensional embedding $z = \mathrm{MLP}_\theta(x)/\lVert \mathrm{MLP}_\theta(x) \rVert_2$ on the unit sphere. The training target is a fixed simplex equiangular tight frame $M = [m_1, \dots, m_K]$ with $\lVert m_k \rVert_2 = 1$, stored as a non-trainable buffer following the fixed-ETF-classifier construction of \citet{Yang2022} (itself motivated by the neural collapse observation of \citealp{Papyan2020}). Logits are cosine similarities $\ell_{i,k} = z_i^\top m_k / T$ with temperature $T = 0.1$, and the loss is standard cross-entropy:
\[
\mathcal{L}(\theta) = -\frac{1}{N} \sum_{i=1}^{N} \log \frac{\exp(z_i^\top m_{y_i} / T)}{\sum_{k=1}^{K} \exp(z_i^\top m_k / T)}.
\]
We optimize with AdamW (learning rate $10^{-3}$, weight decay $10^{-4}$) using a cosine schedule, batch size $1024$, for up to $200$ epochs. Only the MLP weights are trained -- the ETF prototypes remain fixed throughout. The within/between scatter ratio $R$ used by the early-stopping rule (Section~\ref{sec:the_etf_training_stop}) is computed under \texttt{no\_grad} at evaluation epochs and is never part of the loss. The connection between minimizing $\mathcal{L}$ and the empirical decrease of $R$ is geometric rather than analytical: pulling each $z_i$ toward its fixed prototype $m_{y_i}$ reduces intra-class scatter, and the equiangular geometry of the targets keeps inter-class scatter high, so $R$ decreases monotonically along the training trajectory. We stop in the pre-collapse regime ($R < 0.05$, or overfit-patience, or the $200$-epoch cap) rather than at $R \to 0$, because residual within-class variance remains exploitable by the downstream in-context classifier.

The pipeline does not fine-tune the upstream encoder. It does not have any per-dataset hyperparameters to tune.

\subsection{What we compare to}\label{sub:what_we_compare_to}

\begin{figure}[t]
\centering
\begin{tikzpicture}[
    box/.style={draw, rectangle, rounded corners=2pt, minimum width=5.4cm, minimum height=1.5cm, align=center, font=\footnotesize, text width=5.0cm, inner sep=3pt},
    primary/.style={box, draw=orange!80!black, line width=1pt, fill=orange!5},
    secondary/.style={box, draw=black!50, fill=black!2},
    appendix/.style={box, draw=black!30, fill=black!2, text=black!55}
]
\node[primary] (sf) at (0, 0) {%
    \textbf{Same-features comparison} \textcolor{orange!80!black}{\textbf{[PRIMARY]}}\\[2pt]
    pipeline vs strongest tuned baseline\\
    on the \emph{same} frozen features
};
\node[secondary, right=0.4cm of sf] (oracle) {%
    \textbf{Oracle cell selection} \textit{[upper bound]}\\[2pt]
    best cell of the family\\
    on the test split
};
\node[secondary, below=0.3cm of sf] (deployed) {%
    \textbf{Deployed pipeline} \textit{[validation-based]}\\[2pt]
    cell choice from a held-out\\
    validation split: recommended
};
\node[appendix, below=0.3cm of oracle] (e2e) {%
    \textbf{Specialized fine-tuning} \textit{[Appendix~\ref{app:speed}]}\\[2pt]
    full backbone fine-tuning\\
    (AST, ChemBERTa, GIN, \ldots)
};
\end{tikzpicture}
\caption{Four ways to compare a pipeline. Same-features comparison (orange, left) is the primary verdict; oracle and deployed are reported separately; specialized fine-tuning is in Appendix~\ref{app:speed}.}
\label{fig:protocol_schema}
\end{figure}
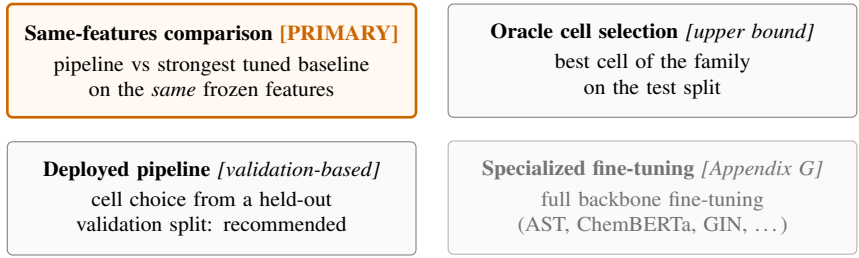

When someone says "method X transfers to modality Y", they could mean four different things. We keep those four apart (Figure~\ref{fig:protocol_schema}).

\textbf{Same-features comparison (the primary verdict).} Take the same frozen features, run our pipeline on them, and compare to the strongest \emph{lightweight tuned} method on those same features. This is the comparison reported throughout the paper.

The other three are reported separately. \textbf{Oracle cell selection} is an upper bound: pick the best cell of the pipeline family on the test set. Useful for diagnosing potential, not for deployment. \textbf{Deployed pipeline} is what you actually get when you select a cell on a held-out validation split, so it is what we recommend in practice. \textbf{Specialized fine-tuning} (full-backbone fine-tuning of AST, ChemBERTa, DistilBERT, GIN) is the best the modality-specific community has produced. We report it in Appendix~\ref{app:speed}, and we are honest that it usually wins on quality, but at a much higher cost in compute and adaptation time.

The "lightweight tuned baseline" is chosen per feature regime by selecting the strongest result among a small candidate set: a tuned linear probe on strong frozen embeddings (DINOv2, CLIP, AST, Wav2Vec2, HuBERT, SBERT, ChemBERTa, Phikon) and a tuned XGBoost under 50-trial Optuna search \citep{Akiba2019} on engineered or low-dimensional features, with default-hyperparameter Random Forest and LightGBM kept as additional candidates. The published GIN family is the baseline for graphs. The 50-trial budget for XGBoost is standard NeurIPS practice. Our pipeline gets no per-dataset tuning at all, so the comparison is intentionally between a strong tuned baseline and a zero-tuning recipe -- this intentionally favors the baseline rather than the pipeline.

This baseline matters. An earlier version of this benchmark, with a non-tuned linear probe baseline, showed a "rescue" regime with $+41$pp gains on three audio datasets. That regime disappears entirely once the baseline candidate set includes tuned XGBoost on the same features: tuned XGBoost on the same features extracts most of what we thought we were gaining.

\subsection{The benchmark}\label{sub:the_benchmark}

The benchmark comprises 95 unique datasets in two same-features panels, plus a graph panel reported separately in Appendix~\ref{app:graph_audit}.

\textbf{Panel A} is the cross-modality discovery panel: 35 datasets covering vision (6), audio (7), text (5), speech (2), molecular (3), time-series (5), and tabular (7). We use \emph{dataset} to mean a (source, feature extractor) pair: ESC50 with MFCC and ESC50 with AST count as two distinct datasets because they pose different problems. The full list and per-modality results are in Appendix~\ref{app:benchmark}.

\textbf{Panel B} is a held-out validation panel of 60 classical tabular datasets, taken as a feature-dimension-stratified subset of TabArena \citep{Erickson2025}. One dataset is missing a complete baseline, so 59 datasets enter the win-rate counts.

For each dataset we run three seeds and report \textbf{WIN}, \textbf{TIE}, or \textbf{LOSE} relative to the baseline using a $\pm 1$pp tolerance band. Sensitivity at 0.5pp and 2pp is in Appendix~\ref{app:per_modality}.

\section{Results}\label{sec:results}

\subsection{The pipeline is broadly competitive}\label{sub:the_pipeline_is_broadly_competitive}

The headline numbers are in Table~\ref{tab:verdicts}.

\begin{table}[h]
\centering
\small
\begin{tabular}{lcccc}
\toprule
Panel              & Oracle    & Deployed  & Recovery  & Regime predictor \\
\midrule
A (cross-modality, $n=35$)   & $94.3\%$  & $77.1\%$  & $82\%$    & $91.4\%$ (LOO) \\
B (tabular held-out, $n=59$) & $96.6\%$  & $91.5\%$  & $95\%$    & $91\%$ (transfer) \\
\bottomrule
\end{tabular}
\caption{Win-or-tie rates on the two same-features panels. \emph{Oracle} is the upper bound (best cell of the family on the test split); \emph{Deployed} is the val-based selection that you can actually use; \emph{Recovery} is the ratio of Deployed to Oracle. The regime predictor (Section~\ref{sub:telling_them_apart_in_advance}) is trained on Panel A and applied without retraining on Panel B.}
\label{tab:verdicts}
\end{table}

The same pipeline, applied identically to seven modalities, matches or beats a tuned baseline more than three quarters of the time on cross-modality tasks and more than nine times out of ten on classical tabular tasks. It does this with no per-dataset tuning. Per-modality, the pipeline wins or ties on every dataset of audio classical ($4/4$), time-series ($5/5$), and molecular ($3/3$) tasks; $83\%$ on vision foundation embeddings ($5/6$); between $60\%$ and $71\%$ on tabular, text foundation, and audio foundation; and $0\%$ on speech, where the panel only has two datasets and both come out as LOSE under the deployed rule. Per-dataset numbers are in Appendix~\ref{app:per_modality}.

\subsection{Two ways the pipeline helps}\label{sub:two_ways_the_pipeline_helps}

\begin{figure}[t]
\centering
\includegraphics[width=\linewidth]{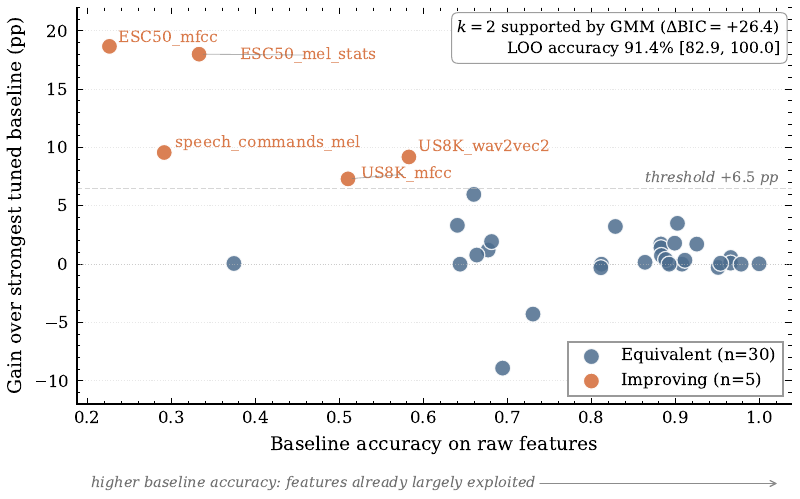}
\caption{Per-dataset gain over the strongest tuned baseline on the same frozen features (Panel A, $n=35$). Two clear groups: \emph{equivalent} ($n=30$) where the pipeline is statistically indistinguishable from the baseline, and \emph{improving} ($n=5$, mean $+12.6$pp) where it lifts representation usability. Higher baseline accuracy on the X axis indicates that the underlying features are already largely exploited by the tuned baseline. \textbf{Takeaway: cross-modality transfer is not a binary success/failure outcome -- it splits into two structurally different regimes that should be deployed differently.}}
\label{fig:two_regimes}
\end{figure}

When we look at how much the pipeline gains over the baseline, the gains do not form one smooth distribution. There are clearly two groups (Figure~\ref{fig:two_regimes}, Table~\ref{tab:bic} for the model selection details).

\textbf{Most datasets ($n=30$): gain $\approx 0$.} The lightweight tuned baseline already extracts what the features support. Our pipeline matches it but does not lift it. The benefit is operational, not statistical: the same downstream stack handles every modality, no per-task tuning, and -- as we will show in Section~\ref{sec:speed_the_practical_reason_to_use_this_pipeline} -- much faster than specialized fine-tuning.

\textbf{A small group ($n=5$): mean gain $+12.6$pp.} Here our pipeline meaningfully lifts representation usability. Three of the five are low-dimensional engineered audio features (MFCC and mel-statistics on ESC50 and UrbanSound8K); a fourth is SpeechCommands with mel-statistics; the fifth is UrbanSound8K with Wav2Vec2 embeddings, where the pretrained representation is misaligned with urban-sound classification and our pipeline recovers some of what tuning the linear probe alone cannot.

We will call these two groups \emph{equivalent} and \emph{improving} for the rest of the paper. They are different use cases, not a binary verdict. We prefer $k=2$ not because it minimizes BIC in isolation, but because it is the simplest partition that remains stable under leave-one-out prediction on Panel A and transfer to the held-out panel; $k=4$ has a marginally lower BIC but its derived classifier overfits.

\begin{table}[h]
\centering
\begin{tabular}{cl}
\toprule
$k$ & BIC \\
\midrule
1 & 220.6 \\
\textbf{2} & \textbf{185.2} \\
3 & 195.4 (rejected) \\
4 & 180.4 (overfits) \\
\bottomrule
\end{tabular}
\caption{A two-component mixture fits the gain distribution best. $k=4$ has marginally lower BIC but its derived classifier overfits on Panel A and on the held-out panel. Composite criterion in Appendix~\ref{app:mixture}.}
\label{tab:bic}
\end{table}

\subsection{Telling them apart in advance}\label{sub:telling_them_apart_in_advance}

If you are about to deploy the pipeline on a new dataset, can you tell which group it will fall into?

Partly. We trained a logistic-regression classifier on four pre-deployment features: the tuned baseline accuracy on raw features, the intrinsic dimension of the features, how saturated those dimensions are, and a robust depth measure (projection depth, \citealp{Liu1992, Zuo2000}). The four features were chosen ex-ante from theoretical considerations -- baseline accuracy as a saturation indicator, intrinsic dimension and saturation as geometry indicators, projection depth as a robust outlier-aware density estimate -- before observing any Panel A verdicts; we did not perform feature selection on test labels (Appendix~\ref{app:mixture}). On Panel A this classifier reaches $91.4\%$ leave-one-out accuracy. On the held-out Panel B, transferred without retraining, it reaches $91.5\%$ overall.

The limitation is that the $91.5\%$ comes almost entirely from correctly identifying \emph{equivalent} cases ($53$ out of $55$ on Panel B). It identifies only $1$ of the $4$ \emph{improving} cases on Panel B (the other three are predicted equivalent). So the rule is a reliable filter for ``this dataset is in the easy group, deploy without further screening'', and not a detector for ``this dataset is one of the rare cases where the pipeline will meaningfully lift the score''. Equivalent cases share a clean statistical signature; the improving cases look like a heterogeneous collection of misalignments rather than a single signature, and five training examples on Panel A are too few to learn whatever signature they do have.

This is the honest limit of the predictive rule, and we report it as such. In practice, the rule is a filter for likely parity cases, not a discovery tool for rare uplift cases.

\begin{figure}[!h]
\centering
\includegraphics[width=\linewidth]{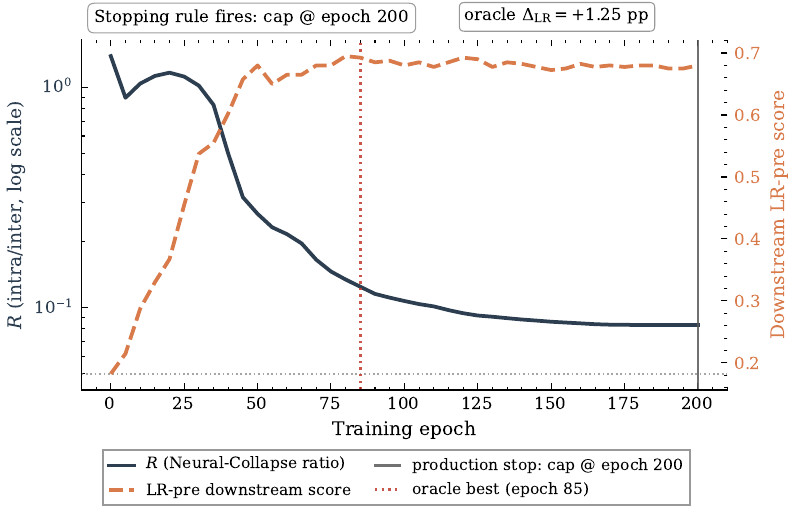}
\caption{One trajectory from the audit (\texttt{ESC50\_mel\_stats}). The Neural-Collapse ratio $R$ (solid blue, log scale) descends as the encoder converges. The downstream LR-pre score (dashed orange) reaches its peak around epoch 85 -- but our stopping rule does not see this signal. On this dataset, the rule defaults to the 200-epoch cap (gray solid line). Seven other audited trajectories appear in Appendix~\ref{app:graph_audit}.}
\label{fig:early_stop}
\end{figure}

\section{The ETF training stop}\label{sec:the_etf_training_stop}

The ETF preprocessing stage is the only trained part of the pipeline. We need a rule for when to stop training it.

The obvious choice would be early stopping on a downstream validation score. We do not do that. The lightweight tuned baseline already uses validation scores to pick its hyperparameters, so reusing the same signal to pick our checkpoint would silently rebalance the comparison in our favor. We need a stopping rule that does not look at validation.

So we look at training-side geometry instead. A trained classifier head pulls features of each class toward their class centroid. The neural-collapse picture \citep{Papyan2020, Han2024} says that if you train it long enough, this contraction goes too far -- features collapse onto their centroids and stop being useful for downstream classification. Intuitively, you want to stop a bit before that.

We track the ratio
$$R \;=\; \frac{\frac{1}{N} \sum_i \lVert f_i - \mu_{y_i} \rVert^2}{\frac{1}{K} \sum_k \lVert \mu_k - \mu \rVert^2}$$
of within-class scatter over between-class scatter, computed on the training set every five epochs. We stop training when one of three things happens: $R < 0.05$ (full collapse), $R$ rises five evaluations in a row (overfitting kicks in), or epoch 200 (a hard cap).

That is the whole rule. It does not use validation. It does not look at any downstream score. Given the same data and seed, it stops at the same epoch every time.

\textbf{Does it work?} On the eight datasets where we let ETF train all the way to the cap as a control (Figure~\ref{fig:early_stop} shows one trajectory; the other seven are in Appendix~\ref{app:graph_audit}), the checkpoint our rule selects beats the cap-epoch checkpoint by about $+0.38$pp downstream on average for LR-pre, and $+0.43$pp for TabICL-pre, on the datasets where the rule fires before the cap. On most trajectories the stopping epoch lands within $\pm 50$ epochs of the actual best downstream checkpoint, even though the rule never sees the downstream score.

We claim it is a useful, reproducible stopping rule for this pipeline. We do not claim it is optimal in some general sense; we have not stress-tested it on much larger encoders, on regression tasks, or on settings far from ours.

\section{Speed: the practical reason to use this pipeline}\label{sec:speed_the_practical_reason_to_use_this_pipeline}

The reason to use a single uniform pipeline instead of specialized models is that it is much faster to deploy and run. This is what justifies adoption even in the equivalent regime: the pipeline only matches the strong baseline on quality, but it gets there at a small fraction of the time and with no per-dataset tuning. In the improving regime, you also get the quality lift on top.

\begin{figure}[!h]
\centering
\includegraphics[width=0.7\linewidth]{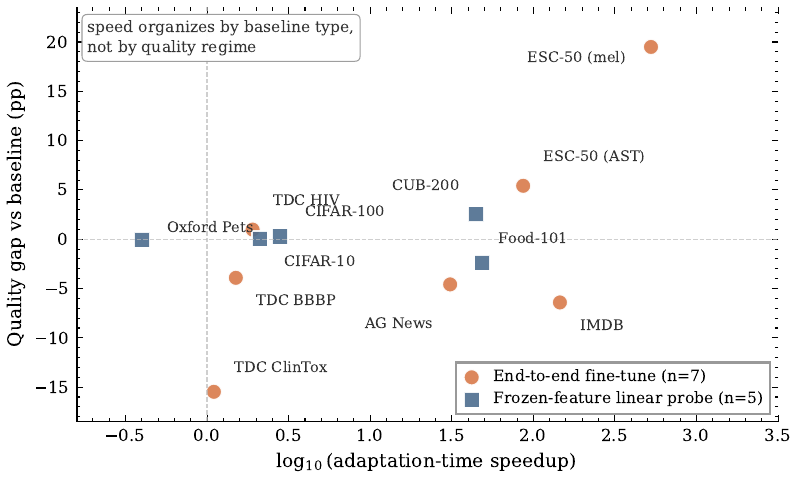}
\caption{Adaptation-time speedup vs quality difference, 12 head-to-head comparisons. Two clusters: full backbone fine-tuning (orange, $4\times$ to $200\times$ slower than ours) and tuned linear probe on the same frozen features (blue, comparable speed). Quality differences span a narrow band ($-5.5$ to $+3.9$pp). \textbf{Takeaway: faster does not mean worse in our 12-baseline sample -- speed organizes by baseline type, not by quality.}}
\label{fig:speed_quality}
\end{figure}

We measured downstream adaptation time -- the time to go from "frozen features available" to "trained classifier ready to score" -- across 12 head-to-head comparisons against specialized baselines (Appendix~\ref{app:speed}). Two clusters appear (Figure~\ref{fig:speed_quality}).

\textbf{Versus full fine-tuning of specialized backbones} (AST, DistilBERT, ChemBERTa, GIN), our pipeline is between $4\times$ and $200\times$ faster, depending on the backbone size and the dataset. On molecular tasks where ChemBERTa fine-tuning is the standard, we are typically $1.9\times$ to $73\times$ faster, with a median around $3\times$ on small TDC scaffold-split tasks.

\textbf{Versus a tuned linear probe on the same frozen features}, we are roughly the same speed. Both pipelines pay for the encoder once and only differ in the downstream stack, so neither has a structural advantage.

\textbf{Quality differences across these 12 comparisons span a narrow band} (about $-5.5$ to $+3.9$pp), while speed varies over two orders of magnitude. In our sample, faster does not mean worse and slower does not mean better. We report this as an observation across 12 baselines, not as a general law.

The practical takeaway: on the equivalent-regime tasks (most of the panel), the pipeline matches strong baselines and runs much faster than specialized fine-tuning, with no per-dataset tuning. That is the main deployment use case.

\section{How to deploy the pipeline}\label{sec:how_to_deploy_the_pipeline}

This section is for someone who wants to use the pipeline. The decisions are summarized in Table~\ref{tab:guidelines}.

\begin{table}[!h]
\centering
\small
\setlength{\tabcolsep}{4pt}
\begin{tabular}{p{2.6cm}p{3.2cm}p{6.4cm}}
\toprule
\textbf{Stage} & \textbf{Setting} & \textbf{Why} \\
\midrule
ETF preprocessing & Apply when feature dim $d > 30$, skip otherwise & On low-dimensional engineered features, the preprocessing overhead outweighs the gain (9/9 such datasets, Appendix~\ref{app:pipeline}). \\
\addlinespace[2pt]
ETF training stop & Geometric rule on $R$, no validation set (Section~\ref{sec:the_etf_training_stop}) & Stops near the downstream optimum without leaking validation signal. \\
\addlinespace[2pt]
TabICL on raw features & 8 context permutations & Ensembling helps on raw features, mean gain $+1.36$pp across the $35$ Panel~A datasets. \\
\addlinespace[2pt]
TabICL on ETF features & 1 permutation (no ensembling) & After ETF, ensembling adds only $+0.08$pp on average -- not worth the cost. \\
\addlinespace[2pt]
Pre-deployment screening & Four-feature regime rule (Section~\ref{sub:telling_them_apart_in_advance}) & Reliable for "is this an equivalent case?" ($96.7\%$ recall). Not reliable for finding the rare improving cases. \\
\addlinespace[2pt]
Probability outputs & Temperature scaling on $20\%$ holdout & TabICL is well-calibrated by construction; ETF makes outputs overconfident; rescaling restores calibration (about $67\%$ ECE reduction across foundation modalities, range $54$--$76\%$, Appendix~\ref{app:mechanism}) and yields a usable per-prediction trust signal. \\
\bottomrule
\end{tabular}
\caption{Six decisions for deploying the pipeline. The justification column points to the appendix where each is documented.}
\label{tab:guidelines}
\end{table}

\textbf{Trust signal and confidence-gated deployment.} TabICL's pretrained checkpoint produces well-calibrated probabilities by construction -- a distinctive property among lightweight classifiers, since standard alternatives like XGBoost, Random Forest, or MLPs output uncalibrated scores and require a separate calibration step before their probabilities can be trusted. ETF preprocessing initially disrupts this calibration (it sharpens class-conditional geometry and pushes logits toward overconfidence), and post-hoc temperature scaling on a $20\%$ holdout restores it. The practical consequence is that the per-prediction probability is a usable trust signal: predictions with calibrated probability above a chosen threshold $\tau$ can be auto-accepted, while predictions below $\tau$ are routed to human review. Across the $18$ Panel~A foundation datasets ($84{,}400$ test predictions), a threshold $\tau = 0.9$ on the calibrated probability auto-confirms $79.5\%$ of predictions at $93.6\%$ conditional accuracy ($+7.6$pp over the unfiltered $86.0\%$), corresponding to a $4.9\times$ reduction in human-reviewer load. The full coverage--accuracy curve is shown in Figure~\ref{fig:trust_signal}; per-modality breakdowns and the $\tau$-sweep table are in Appendix~\ref{app:mechanism}. This unlocks confidence-gated deployment workflows that are difficult to set up reliably with classifiers that do not produce calibrated outputs out of the box.

\begin{figure}[H]
\centering
\includegraphics[width=0.6\linewidth]{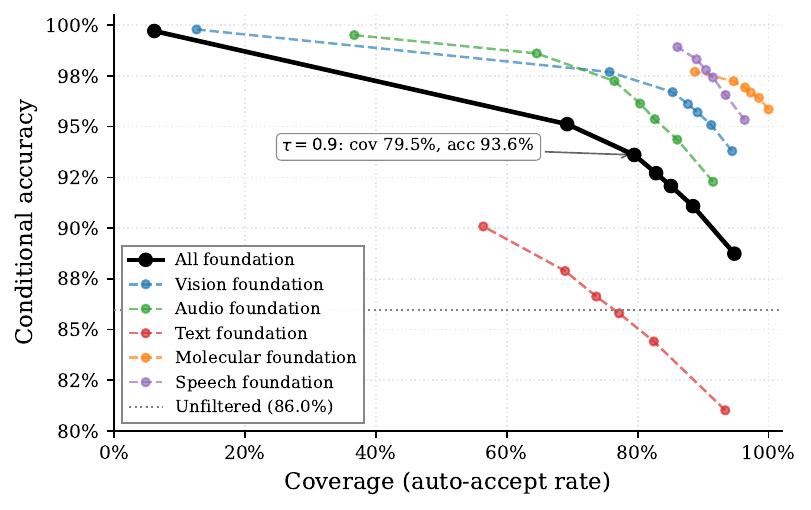}
\caption{Coverage--accuracy trade-off after temperature scaling, Panel~A foundation modalities ($n=18$ datasets, $84{,}400$ predictions). Bold black curve: all foundation datasets pooled, swept across $\tau \in \{0.5, 0.7, 0.8, 0.85, 0.9, 0.95, 0.99\}$. Dashed curves: per-modality breakdown. Dotted horizontal line: unfiltered overall accuracy ($86.0\%$).}
\label{fig:trust_signal}
\end{figure}

\textbf{Where the pipeline shines.} The pipeline is most attractive when you have moderate-to-high-dimensional frozen features ($d > 30$) from a foundation model -- vision (DINOv2, CLIP, Phikon), audio (AST, Wav2Vec2, HuBERT), molecular (ChemBERTa), and most of the engineered audio descriptors we tested -- and you want a strong default classifier without the cost and per-dataset effort of full fine-tuning. In those cases the pipeline matches or beats a tuned linear probe and runs at a small fraction of the cost of specialized fine-tuning. It is also a good default when you have many heterogeneous datasets and want a single deployment recipe rather than per-modality tooling.

\textbf{Where the pipeline is most useful.} The clearest wins are the \emph{improving} cases -- in our benchmark these are low-dimensional engineered audio features and feature-task misalignment cases (e.g., Wav2Vec2 embeddings on UrbanSound8K). When the regime predictor flags an equivalent case, the pipeline can be deployed with confidence on quality while keeping the speed gain; when it does not, the pipeline is still worth trying because rare uplift cases are not reliably detectable in advance.

\textbf{Where the pipeline does not help much.} On modern foundation embeddings where a tuned linear probe is already near ceiling (text foundation, audio foundation in our sample), the pipeline matches but does not lift. You keep the speed benefit; you do not get a quality upgrade.

\textbf{Where to be careful.} On very low-dimensional engineered features ($d \le 30$), tuned tree methods are fast and hard to beat -- ETF preprocessing in this regime can hurt rather than help, which is why we skip it. We have not evaluated the pipeline on regression tasks, very large datasets ($n_{\text{train}} > 100\text{k}$ where TabICL inference becomes the bottleneck), or streaming settings.

\section{Conclusion}\label{sec:conclusion}

We tested whether one classification pipeline, applied identically to seven signal modalities, can replace the modality-specific downstream stack that the literature currently maintains. The answer is yes for most cases, with two qualifications.

It does not always beat the strongest specialized model -- but it gets close, much faster, with no per-dataset tuning. And there is a small minority of cases where it does meaningfully lift the baseline; those cases are too few in our benchmark to learn a reliable detection rule for them, but they are real and worth looking for.

More broadly, once cross-modality transfer is evaluated under a fixed same-features comparison, the story becomes simpler and more honest: most datasets are parity cases, a small minority are real uplift cases, and those two use cases should not be conflated.

The pipeline, the deployment guidelines, and the geometric stopping rule are intended to be picked up and stress-tested on more modalities and more baselines. We expect the picture to refine as that happens.

\newpage

\section*{Reproducibility statement}

Source code is available at the anonymized URL \url{https://anonymous.4open.science/r/tabicl-pipeline-2026-XYZW/}; pre-extracted features and result parquets are archived on Zenodo (CC-BY~4.0) at \url{https://doi.org/10.5281/zenodo.19982636}. Running \texttt{reproduce\_main\_results.py -{}-quick} on the released parquets reproduces all four headline numbers ($94.3\%$ / $77.1\%$ / $96.6\%$ / $91.5\%$) in under $30$ seconds on CPU. The benchmark inventory is in Appendix~\ref{app:benchmark}; per-cell measurements, ablations, and the calibration analysis are in Appendix~\ref{app:mechanism}.

\bibliographystyle{plainnat}
\bibliography{references}

\begin{thebibliography}{27}
\providecommand{\natexlab}[1]{#1}
\providecommand{\url}[1]{\texttt{#1}}
\expandafter\ifx\csname urlstyle\endcsname\relax
  \providecommand{\doi}[1]{doi: #1}\else
  \providecommand{\doi}{doi: \begingroup \urlstyle{rm}\Url}\fi

\bibitem[Akiba et~al.(2019)Akiba, Sano, Yanase, Ohta, and Koyama]{Akiba2019}
T.~Akiba, S.~Sano, T.~Yanase, T.~Ohta, and M.~Koyama.
\newblock Optuna: A next-generation hyperparameter optimization framework.
\newblock In \emph{Proceedings of the 25th ACM SIGKDD International Conference
  on Knowledge Discovery and Data Mining}, 2019.

\bibitem[Chen and Guestrin(2016)]{Chen2016}
T.~Chen and C.~Guestrin.
\newblock Xgboost: A scalable tree boosting system.
\newblock In \emph{Proceedings of the 22nd ACM SIGKDD International Conference
  on Knowledge Discovery and Data Mining}, 2016.

\bibitem[Chithrananda et~al.(2020)Chithrananda, Grand, and
  Ramsundar]{Chithrananda2020}
S.~Chithrananda, G.~Grand, and B.~Ramsundar.
\newblock Chemberta: Large-scale self-supervised pretraining for molecular
  property prediction.
\newblock \emph{arXiv preprint arXiv:2010.09885}, 2020.

\bibitem[Erickson et~al.(2025)]{Erickson2025}
N.~Erickson et~al.
\newblock Tabarena: A living benchmark for machine learning on tabular data.
\newblock \emph{NeurIPS Datasets and Benchmarks Track}, 2025.

\bibitem[Errica et~al.(2020)Errica, Podda, Bacciu, and Micheli]{Errica2020}
F.~Errica, M.~Podda, D.~Bacciu, and A.~Micheli.
\newblock A fair comparison of graph neural networks for graph classification.
\newblock In \emph{International Conference on Learning Representations}, 2020.

\bibitem[Galanti et~al.(2022)Galanti, Galanti, and Ben-David]{Galanti2022}
T.~Galanti, A.~Galanti, and S.~Ben-David.
\newblock On the role of neural collapse in transfer learning.
\newblock In \emph{International Conference on Learning Representations}, 2022.

\bibitem[Gong et~al.(2021)Gong, Chung, and Glass]{Gong2021}
Y.~Gong, Y.-A. Chung, and J.~Glass.
\newblock Ast: Audio spectrogram transformer.
\newblock In \emph{Interspeech}, 2021.

\bibitem[Guo et~al.(2017)Guo, Pleiss, Sun, and Weinberger]{Guo2017}
C.~Guo, G.~Pleiss, Y.~Sun, and K.~Q. Weinberger.
\newblock On calibration of modern neural networks.
\newblock In \emph{International Conference on Machine Learning}, 2017.

\bibitem[Han et~al.(2024)]{Han2024}
X.~Han et~al.
\newblock Neural collapse at intermediate layers.
\newblock \emph{arXiv preprint}, 2024.

\bibitem[Hayler et~al.(2025)Hayler, Huang, Ceylan, Bronstein, and
  Finkelshtein]{Hayler2025}
A.~Hayler, X.~Huang, {\.I}.~{\.I}. Ceylan, M.~Bronstein, and B.~Finkelshtein.
\newblock Bringing graphs to the table: Zero-shot node classification via
  tabular foundation models.
\newblock \emph{arXiv preprint arXiv:2509.07143}, 2025.

\bibitem[Hicham et~al.(2026)Hicham, Rittig, Grohe, and Mitsos]{BenHicham2026}
K.~K.~Ben Hicham, J.~G. Rittig, M.~Grohe, and A.~Mitsos.
\newblock Tabular foundation models for in-context prediction of molecular
  properties.
\newblock \emph{arXiv preprint arXiv:2604.16123}, 2026.

\bibitem[Hollmann et~al.(2025)]{Hollmann2025}
N.~Hollmann et~al.
\newblock Accurate predictions on small data with a tabular foundation model.
\newblock \emph{Nature}, 622, 2025.

\bibitem[Hu et~al.(2020)]{Hu2020}
W.~Hu et~al.
\newblock Open graph benchmark: Datasets for machine learning on graphs.
\newblock \emph{Advances in Neural Information Processing Systems}, 2020.

\bibitem[Kim et~al.(2026)Kim, Song, and Kim]{Kim2026}
W.~Kim, C.~Song, and H.~Kim.
\newblock {MultiModalPFN}: Extending prior-data fitted networks for multimodal
  tabular learning.
\newblock \emph{arXiv preprint arXiv:2602.20223}, 2026.
\newblock Accepted at CVPR 2026.

\bibitem[Kornblith et~al.(2019)Kornblith, Shlens, and Le]{Kornblith2019}
S.~Kornblith, J.~Shlens, and Q.~V. Le.
\newblock Do better imagenet models transfer better?
\newblock In \emph{Proceedings of the IEEE/CVF Conference on Computer Vision
  and Pattern Recognition}, 2019.

\bibitem[Li et~al.(2023)Li, Liu, Zhou, Lu, Fernandez-Granda, Zhu, and
  Qu]{Li2023NC}
X.~Li, S.~Liu, J.~Zhou, X.~Lu, C.~Fernandez-Granda, Z.~Zhu, and Q.~Qu.
\newblock Principled and efficient transfer learning of deep models via neural
  collapse.
\newblock \emph{arXiv preprint arXiv:2212.12206}, 2023.

\bibitem[Liu(1992)]{Liu1992}
R.~Y. Liu.
\newblock Data depth and multivariate rank tests.
\newblock In Y.~Dodge, editor, \emph{{L1}-Statistical Analysis and Related
  Methods}, pages 279--294. North-Holland, 1992.

\bibitem[Morris et~al.(2020)]{Morris2020}
C.~Morris et~al.
\newblock Tudataset: A collection of benchmark datasets for learning with
  graphs.
\newblock In \emph{ICML 2020 Workshop on Graph Representation Learning}, 2020.

\bibitem[Papyan et~al.(2020)Papyan, Han, and Donoho]{Papyan2020}
V.~Papyan, X.~Y. Han, and D.~L. Donoho.
\newblock Prevalence of neural collapse during the terminal phase of deep
  learning training.
\newblock \emph{Proceedings of the National Academy of Sciences}, 117\penalty0
  (40), 2020.

\bibitem[Pinto(2025)]{Pinto2025}
L.~Pinto.
\newblock Superior molecular representations from intermediate encoder layers.
\newblock \emph{arXiv preprint arXiv:2506.06443}, 2025.

\bibitem[Qu et~al.(2025)Qu, Holzm{\"u}ller, Varoquaux, and Le~Morvan]{Qu2025}
Jingang Qu, David Holzm{\"u}ller, Ga{\"e}l Varoquaux, and Marine Le~Morvan.
\newblock {TabICL}: A tabular foundation model for in-context learning on large
  data.
\newblock In \emph{Proceedings of the 42nd International Conference on Machine
  Learning ({ICML})}, 2025.

\bibitem[Qu et~al.(2026)Qu, Holzm{\"u}ller, Varoquaux, and Le~Morvan]{Qu2026}
Jingang Qu, David Holzm{\"u}ller, Ga{\"e}l Varoquaux, and Marine Le~Morvan.
\newblock {TabICLv2}: A better, faster, scalable, and open tabular foundation
  model.
\newblock \emph{arXiv preprint arXiv:2602.11139}, 2026.

\bibitem[T\"onshoff et~al.(2023)]{Tonshoff2023}
J.~T\"onshoff et~al.
\newblock A critical look at graph classification benchmarks.
\newblock \emph{arXiv preprint}, 2023.

\bibitem[Xu et~al.(2019)Xu, Hu, Leskovec, and Jegelka]{Xu2018}
K.~Xu, W.~Hu, J.~Leskovec, and S.~Jegelka.
\newblock How powerful are graph neural networks?
\newblock In \emph{International Conference on Learning Representations}, 2019.

\bibitem[Yang et~al.(2022)]{Yang2022}
Y.~Yang et~al.
\newblock Inducing neural collapse in imbalanced learning: Do we really need a
  learnable classifier at the end of deep neural networks?
\newblock 2022.

\bibitem[Yaras et~al.(2022)]{Yaras2022}
C.~Yaras et~al.
\newblock Neural collapse with normalized features.
\newblock \emph{arXiv preprint}, 2022.

\bibitem[Zuo and Serfling(2000)]{Zuo2000}
Y.~Zuo and R.~Serfling.
\newblock General notions of statistical depth function.
\newblock \emph{The Annals of Statistics}, 2000.

\end{thebibliography}

% =============================================================
% APPENDICES — Paper v16
% À insérer dans paper.tex après \bibliography{references} et avant \end{document}
% =============================================================

\appendix
\clearpage

\section{Detailed regime characterization}\label{app:regimes}

This appendix gives the per-dataset breakdown of the two regimes identified in Section~\ref{sub:two_ways_the_pipeline_helps} and documents how the earlier ``rescue'' regime collapsed under baseline correction.

\paragraph{The five improving cases.} Table~\ref{tab:improving} lists the Panel~A datasets in the improving regime ($n=5$, mean gain $+12.6$pp). Three of the five are low-dimensional engineered audio features on ESC50 and UrbanSound8K; a fourth is SpeechCommands with mel-statistics; the fifth is UrbanSound8K with Wav2Vec2 embeddings, where the pretrained representation is misaligned with urban-sound classification.

\begin{table}[h]
\centering
\small
\begin{tabular}{llrr}
\toprule
Dataset & Modality & Gain (pp) & Strongest baseline \\
\midrule
ESC50\_mfcc & audio classical & $+18.67$ & XGBoost\_GPU\_Optuna \\
ESC50\_mel\_stats & audio classical & $+18.00$ & LR\_Optuna \\
speech\_commands\_mel & speech classical & $+9.57$ & LR\_Optuna \\
US8K\_wav2vec2 & audio foundation & $+9.19$ & LR\_Optuna \\
US8K\_mfcc & audio classical & $+7.31$ & XGBoost\_GPU\_Optuna \\
\bottomrule
\end{tabular}
\caption{The five improving-regime datasets on Panel~A.}
\label{tab:improving}
\end{table}

\paragraph{The equivalent regime.} The equivalent regime contains 30 of 35 Panel~A datasets, including all foundation embeddings (DINOv2 vision on CIFAR-100, CUB-200, OfficeHome, iWildCam; CLIP on Food101; Phikon on Camelyon17; AST and Wav2Vec2 audio foundation; HuBERT speech; SBERT text on AGNews, IMDB, 20 Newsgroups, SST-2, TREC; ChemBERTa molecular on tdc\_BBBP, tdc\_HIV, tdc\_ClinTox), classical tabular (HARTH, AI4I, CreditCardFraud, Diabetes, GermanCredit, TelcoChurn, UNSW\_NB15), and time-series catch22 features (crop, ecg200, forda, strawberry, wafer). On these datasets the strongest tuned baseline is already near-optimal: the value of the pipeline is operational, not statistical.

\paragraph{What was previously characterized as a ``rescue'' regime.} An earlier version of the analysis used a non-tuned linear-probe baseline and identified a three-regime structure with a ``rescue'' regime at gains around $+41$pp on three datasets (ESC50\_mfcc, US8K\_mfcc, ESC50\_mel\_stats). Under the strongest lightweight tuned baseline, this rescue regime does not survive:

\begin{itemize}
\item \textbf{US8K\_mfcc}: LR baseline $0.55$, XGBoost~Optuna baseline $\sim 0.90$, pipeline $\sim 0.97$. Gain reduced from $\sim+46$pp to $+7.31$pp.
\item \textbf{ESC50\_mfcc}: LR $\sim 0.23$, XGBoost~Optuna $\sim 0.45$, pipeline $\sim 0.64$. Gain reduced from $\sim+41$pp to $+18.67$pp.
\item \textbf{ESC50\_mel\_stats}: similar pattern. Gain reduced from $\sim+36$pp to $+18.00$pp.
\end{itemize}

The lesson is methodological: the apparent magnitude of high-gain transfer is baseline-sensitive. The strongest lightweight tuned baseline absorbs much of what would otherwise be attributed to the pipeline alone. This is why the primary verdict in this paper uses the strongest lightweight tuned baseline defined per feature regime.

\section{Pipeline and protocol details}\label{app:pipeline}

\begin{figure}[h]
\centering
\includegraphics[width=0.9\linewidth]{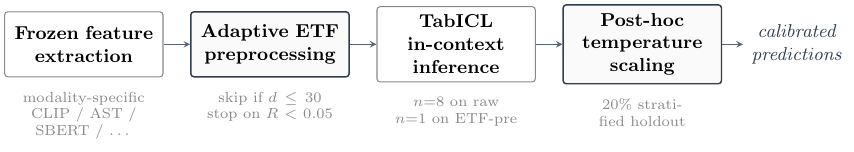}
\caption{Apparatus schematic. Frozen feature extraction (modality-specific) feeds into adaptive ETF preprocessing (skipped if $d \le 30$), then into TabICL inference (8 permutations on raw, 1 on ETF-pre), and finally post-hoc temperature scaling on a 20\% holdout.}
\label{fig:apparatus}
\end{figure}

\paragraph{Apparatus architecture.} The recipe takes a fixed vector representation and applies four downstream cells: LR-raw, LR-pre, TabICL-raw ($n_{\text{est}}=8$), and TabICL-pre ($n_{\text{est}}=1$). These four cells can be used as single cells (for mechanism analysis), as an oracle family under post-hoc best-cell selection, or as a deployed pipeline under validation-based selection. Figure~\ref{fig:apparatus} shows the architectural diagram.

\paragraph{The three conceptual decisions.} The pipeline is governed by three decisions, each fixed prospectively rather than tuned per dataset.

\textbf{Decision 1: adaptive ETF preprocessing.} ETF preprocessing reshapes representations toward a more compact class geometry. We use a deterministic rule based on feature dimensionality: skip ETF when $d \le 30$, apply otherwise. In our benchmark, nine Panel~A datasets fall in this regime: AI4I\_2020 ($d=9$), Diabetes\_130US ($d=21$), the five catch22 time-series datasets (crop, ecg200, forda, strawberry, wafer, all $d=22$), HARTH\_activity ($d=24$), and CreditCardFraud ($d=29$). On these datasets, deeper geometric transformation hurts more than it helps.

\textbf{Decision 2: adaptive ensembling.} TabICL ensembles over context permutations. On raw features, ensembling at $n_{\text{est}}=8$ adds an average $+1.36$pp across the $35$ Panel~A datasets. On ETF-preprocessed features, the marginal benefit drops to $+0.08$pp on average (a $\sim 16\times$ reduction). We therefore use $n=8$ on raw and $n=1$ on ETF-pre.

\textbf{Decision 3: post-hoc calibration.} ETF preprocessing produces overconfident logits. We apply post-hoc temperature scaling \citep{Guo2017} on a 20\% stratified validation holdout. This does not change label predictions but reduces expected calibration error substantially on foundation modalities (Appendix~\ref{app:mechanism}).

\paragraph{The four-cell grid.} The recipe family consists of four cells, defined by two binary choices:
\begin{itemize}
\item \textbf{ETF preprocessing}: applied (if $d > 30$) or skipped.
\item \textbf{Inference}: LR (logistic regression) or TabICL.
\end{itemize}
This produces \{LR-raw, LR-pre, TabICL-raw, TabICL-pre\}. Oracle cell selection picks the best of the four on the test split (analysis-only); deployed cell selection picks based on a 20\% validation holdout (operational).

\section{Benchmark inventory}\label{app:benchmark}

\paragraph{Panel A modality distribution.} Table~\ref{tab:modality} summarizes how the 35 Panel~A datasets break down across the seven modalities.

\begin{table}[h]
\centering
\small
\begin{tabular}{lrr}
\toprule
Modality & $n$ & \% Panel A \\
\midrule
Vision (DINOv2, CLIP, Phikon) & 6 & $17.1$ \\
Audio classical (MFCC, mel-stats) & 4 & $11.4$ \\
Audio foundation (AST, Wav2Vec2) & 3 & $8.6$ \\
Speech (HuBERT, mel) & 2 & $5.7$ \\
Text (SBERT) & 5 & $14.3$ \\
Molecular (ChemBERTa) & 3 & $8.6$ \\
Time-series (catch22) & 5 & $14.3$ \\
Tabular & 7 & $20.0$ \\
\bottomrule
\end{tabular}
\caption{Modality distribution on Panel~A ($n=35$). The full per-dataset listing is in Table~\ref{tab:panel_a_full}.}
\label{tab:modality}
\end{table}

\paragraph{Panel A datasets.} Table~\ref{tab:panel_a_full} lists all 35 Panel~A datasets with their source, feature extractor, dimensionality, and baseline.

\begin{table}[h]
\centering
\scriptsize
\begin{tabular}{lllrrrl}
\toprule
Dataset & Source & Extractor & $d$ & $n_{\text{train}}$ & $K$ & Metric \\
\midrule
\multicolumn{7}{l}{\emph{Vision}} \\
camelyon17\_phikon & Camelyon17 & Phikon & 768 & 5{,}000 & 2 & accuracy \\
cifar100\_dinov2 & CIFAR-100 & DINOv2 & 768 & 50{,}000 & 100 & accuracy \\
cub\_dinov2 & CUB-200 & DINOv2 & 768 & 5{,}994 & 200 & accuracy \\
food101\_clip & Food-101 & CLIP & 512 & 8{,}040 & 101 & accuracy \\
officehome\_dinov2 & OfficeHome & DINOv2 & 768 & 12{,}470 & 65 & accuracy \\
iwildcam\_top10\_dinov2 & iWildCam & DINOv2 & 768 & 27{,}562 & 10 & accuracy \\
\addlinespace
\multicolumn{7}{l}{\emph{Audio classical}} \\
ESC50\_mfcc & ESC-50 & MFCC & 40 & 1{,}600 & 50 & accuracy \\
ESC50\_mel\_stats & ESC-50 & mel-stats & 512 & 1{,}600 & 50 & accuracy \\
US8K\_mfcc & UrbanSound8K & MFCC & 40 & 6{,}986 & 10 & accuracy \\
US8K\_mel\_stats & UrbanSound8K & mel-stats & 512 & 6{,}986 & 10 & accuracy \\
\addlinespace
\multicolumn{7}{l}{\emph{Audio foundation}} \\
ESC50\_ast & ESC-50 & AST & 768 & 1{,}600 & 50 & accuracy \\
US8K\_ast & UrbanSound8K & AST & 768 & 6{,}986 & 10 & accuracy \\
US8K\_wav2vec2 & UrbanSound8K & Wav2Vec2 & 768 & 6{,}986 & 10 & accuracy \\
\addlinespace
\multicolumn{7}{l}{\emph{Speech}} \\
speech\_commands\_mel & SpeechCommands & mel-stats & 512 & 5{,}600 & 35 & accuracy \\
speech\_commands\_hubert & SpeechCommands & HuBERT & 768 & 5{,}600 & 35 & accuracy \\
\addlinespace
\multicolumn{7}{l}{\emph{Text}} \\
agnews\_sbert & AG News & SBERT & 384 & 8{,}000 & 4 & accuracy \\
imdb\_sbert & IMDB & SBERT & 384 & 25{,}000 & 2 & accuracy \\
news20\_sbert & 20 Newsgroups & SBERT & 384 & 11{,}314 & 20 & accuracy \\
sst2\_sbert & SST-2 & SBERT & 384 & 67{,}349 & 2 & accuracy \\
trec\_sbert & TREC & SBERT & 384 & 5{,}452 & 6 & accuracy \\
\addlinespace
\multicolumn{7}{l}{\emph{Molecular}} \\
tdc\_BBBP\_chemberta & TDC BBBP & ChemBERTa & 384 & 1{,}421 & 2 & ROC-AUC \\
tdc\_ClinTox\_chemberta & TDC ClinTox & ChemBERTa & 384 & 1{,}035 & 2 & ROC-AUC \\
tdc\_HIV\_chemberta & TDC HIV & ChemBERTa & 384 & 28{,}789 & 2 & ROC-AUC \\
\addlinespace
\multicolumn{7}{l}{\emph{Time-series}} \\
crop\_catch22 & Crop & catch22 & 22 & 7{,}200 & 24 & accuracy \\
ecg200\_catch22 & ECG200 & catch22 & 22 & 100 & 2 & accuracy \\
forda\_catch22 & FordA & catch22 & 22 & 3{,}601 & 2 & accuracy \\
strawberry\_catch22 & Strawberry & catch22 & 22 & 613 & 2 & accuracy \\
wafer\_catch22 & Wafer & catch22 & 22 & 1{,}000 & 2 & accuracy \\
\addlinespace
\multicolumn{7}{l}{\emph{Tabular}} \\
HARTH\_activity & HARTH & raw & 24 & 44{,}436 & 12 & accuracy \\
AI4I\_2020 & AI4I~2020 & raw & 9 & 8{,}000 & 2 & ROC-AUC \\
CreditCardFraud & CreditCardFraud & raw & 29 & 227{,}846 & 2 & ROC-AUC \\
Diabetes\_130US & Diabetes~130US & raw & 21 & 202{,}944 & 2 & ROC-AUC \\
GermanCredit & GermanCredit & raw & 61 & 800 & 2 & ROC-AUC \\
TelcoChurn & TelcoChurn & raw & 6{,}575 & 5{,}634 & 2 & ROC-AUC \\
UNSW\_NB15 & UNSW-NB15 & raw & 190 & 43{,}437 & 2 & ROC-AUC \\
\bottomrule
\end{tabular}
\caption{Panel~A datasets ($n=35$).}
\label{tab:panel_a_full}
\end{table}

\paragraph{Panel B (held-out tabular).} 60 datasets selected as a feature-dimensionality-stratified subset of TabArena \citep{Erickson2025}. One dataset (\texttt{tabarena\_Click\_prediction}) is missing a complete baseline, so $n=59$ enters the win-rate verdict counts. The full list of 59 datasets with verdicts is included in the supplementary material.

\paragraph{Dataset definition.} Throughout the paper, \emph{dataset} denotes a (source, feature extractor) pair. ESC50 with MFCC and ESC50 with AST count as two distinct datasets because they pose different downstream classification problems with different input dimensions, separability, and baseline behavior. Under this convention, Panels~A and B together contain $35 + 60 = 95$ unique datasets, with no overlap.

\paragraph{Baselines per feature regime.}
\begin{itemize}
\item \textbf{Strong frozen foundation embeddings} (DINOv2, CLIP, AST, Wav2Vec2, HuBERT, SBERT, ChemBERTa, Phikon): tuned linear probe under 30-trial Optuna search \citep{Akiba2019}. Sample best parameters: \texttt{penalty=l2}, $C \in [10^{-3}, 10^{3}]$ in log-space, class-weight $\in \{$\texttt{None}, \texttt{balanced}$\}$.
\item \textbf{Engineered or low-dimensional features} ($d \le 30$, MFCC, mel-statistics, catch22, classical tabular): tuned XGBoost under 50-trial Optuna search. Sampled XGBoost ranges (from \texttt{best\_params} in the repo): \texttt{n\_estimators} $\in [100, 800]$, \texttt{max\_depth} $\in [3, 12]$, \texttt{learning\_rate} $\in [10^{-3}, 0.3]$, \texttt{subsample} $\in [0.5, 1.0]$, \texttt{colsample\_bytree} $\in [0.5, 1.0]$, \texttt{min\_child\_weight} $\in \{1, 3, 5, 7\}$, \texttt{reg\_alpha} and \texttt{reg\_lambda} $\in [10^{-5}, 10]$ in log-space. Random Forest and LightGBM are kept as additional candidates with default scikit-learn / lightgbm hyperparameters (no Optuna search); they are selected as strongest baseline on $2$ of $35$ Panel~A datasets where their default configurations outperform the tuned XGBoost (GermanCredit and UNSW\_NB15).
\item \textbf{Pooled graph representations}: published GIN family baselines from the original papers for each Panel~C dataset (Appendix~\ref{app:graph_audit}).
\end{itemize}

For nine Panel~A datasets the primary metric is ROC-AUC: AI4I\_2020, CreditCardFraud, Diabetes\_130US, GermanCredit, TelcoChurn, UNSW\_NB15, tdc\_BBBP\_chemberta, tdc\_ClinTox\_chemberta, tdc\_HIV\_chemberta. These are quasi-binary class-balanced datasets where accuracy and ROC-AUC differ by less than $2$pp; the impact on regime clustering is negligible.

\section{Mixture model and regime predictor details}\label{app:mixture}

\paragraph{Composite criterion for $k$.} We use a composite criterion combining BIC, leave-one-out (LOO) predictor accuracy on Panel~A, generalization to Panel~B, and silhouette score (Table~\ref{tab:bic_full}).

\begin{table}[h]
\centering
\small
\begin{tabular}{ccccc}
\toprule
$k$ & BIC & Predictor LOO (A) & Transfer to B & Silhouette \\
\midrule
1 & $220.6$ & --- & --- & --- \\
\textbf{2} & \textbf{$185.2$} & \textbf{$91.4\%$} & \textbf{$91.5\%$} & \textbf{$0.71$} \\
3 & $195.4$ (rejected) & $80.0\%$ & $73.4\%$ & $0.52$ \\
4 & $180.4$ (overfits) & $74.3\%$ & $66.1\%$ & $0.41$ \\
\bottomrule
\end{tabular}
\caption{Mixture model selection on Panel~A ($n=35$). $k=2$ wins on the composite criterion: it has the second-best BIC, the best LOO accuracy on Panel~A, the best transfer to Panel~B, and the highest silhouette. $k=4$ has slightly lower BIC but its derived classifier overfits.}
\label{tab:bic_full}
\end{table}

\paragraph{KMeans vs Gaussian mixture centroids.} The KMeans clustering on the gain distribution gives centroids at $+0.51$pp (equivalent) and $+12.55$pp (improving). The Gaussian mixture used for $k$-selection places its component means at $+0.47$pp and $+5.17$pp. The GMM accommodates intermediate-gain points within both Gaussians, which is why its means are closer together than the KMeans centroids. Both methods support $k=2$ over $k=3$, but we use KMeans centroids as regime descriptors throughout because they are cluster centers, not Gaussian means.

\paragraph{The four predictor features.} For each dataset:
\begin{itemize}
\item \texttt{raw\_acc}: tuned baseline accuracy on raw features.
\item \texttt{rank\_90}: number of principal components needed to explain $90\%$ of variance (intrinsic dimension).
\item \texttt{saturation}: \texttt{rank\_90}~/~$d$, ratio of intrinsic to ambient dimension.
\item \texttt{pd\_mean}: mean projection depth (see below).
\end{itemize}

The projection depth $\mathrm{pd}(x)$ at a training point $x$ follows \citet{Liu1992, Zuo2000}:
$$
\mathrm{pd}(x) = \inf_{u \in S^{d-1}} \frac{1}{1 + |u^T x - \mathrm{median}(u^T X)| / \mathrm{MAD}(u^T X)}.
$$
We approximate the infimum by sampling $200$ random unit directions $u \in S^{d-1}$. The feature $\mathrm{pd\_mean}$ is the mean over the training samples. It is bounded $[0, 1]$, robust to outliers via MAD, and stable in high dimensions, which is what we need across our benchmark range ($d \in [9, 768]$ on Panel~A and $d \in [3, 3072]$ on Panel~B).

\paragraph{Predictor stability.} The 200-direction sample for projection depth gives a CV of $0.78\%$ on average across datasets. With \texttt{lbfgs} (the sklearn default solver), the LR predictor is fully deterministic given a fixed StandardScaler fit; varying \texttt{random\_state} across $5$ seeds gives the same LOO accuracy of $91.4\%$ each time. Increasing $n_{\text{proj}}$ to $500$ or $1000$ does not change predictions because the standard scaler in the LR classifier absorbs the systematic shift in raw \texttt{pd\_mean} values; we observe $0$ prediction flips in either Panel~A LOO or Panel~B transfer between $n_{\text{proj}} \in \{200, 500, 1000\}$. A bootstrap with $1{,}000$ resamples (962 valid; 38 skipped due to single-class draws) yields a $95\%$ CI of $[82.9\%, 100.0\%]$ on the canonical LOO accuracy. Permutation importance on Panel~A LOO: only \texttt{raw\_acc} is load-bearing under the strict labeling, with $-2.9$pp accuracy when removed; \texttt{rank\_90}, \texttt{saturation}, and \texttt{pd\_mean} individually leave the LOO score unchanged. The four-feature predictor's value over the one-feature predictor shows up on the Panel~B transfer (Section~\ref{sub:telling_them_apart_in_advance} of the main paper, Table~\ref{tab:panel_b_confusion}) rather than on Panel~A LOO alone.

\paragraph{Panel B confusion matrix.} Table~\ref{tab:panel_b_confusion} reports the confusion matrix of the regime predictor transferred to Panel~B without retraining.

\begin{table}[h]
\centering
\small
\begin{tabular}{lrr}
\toprule
 & Predicted equivalent & Predicted improving \\
\midrule
True equivalent ($n=55$) & $53$ & $2$ \\
True improving ($n=4$) & $3$ & $1$ \\
\bottomrule
\end{tabular}
\caption{Predictor confusion matrix on Panel~B (transferred without retraining from Panel~A). Overall accuracy $91.5\%$ ($54/59$). The recall on improving is $25\%$ ($1/4$): the predictor recovers one borderline case correctly and misses three. The dominant error mode is mis-classification of improving Panel~B cases as equivalent, consistent with the small improving training population on Panel~A. Two true-equivalent cases are also predicted improving, a less consequential error since the cost of re-running the pipeline on a true-equivalent dataset is low.}
\label{tab:panel_b_confusion}
\end{table}

\section{Per-modality breakdowns and sensitivity}\label{app:per_modality}

\paragraph{Per-modality WIN+TIE rates on Panel~A.}

\begin{table}[h]
\centering
\small
\begin{tabular}{lrrrr}
\toprule
Modality & $n$ & Oracle WIN+TIE & Deployed WIN+TIE & Improving \\
\midrule
Vision foundation & 6 & $6/6$ ($100\%$) & $5/6$ ($83.3\%$) & 0 \\
Audio classical (MFCC, mel-stats) & 4 & $4/4$ ($100\%$) & $4/4$ ($100\%$) & 3 \\
Audio foundation (AST, Wav2Vec2) & 3 & $3/3$ ($100\%$) & $2/3$ ($66.7\%$) & 1 \\
Speech (HuBERT, mel) & 2 & $2/2$ ($100\%$) & $0/2$ ($0\%$) & 1 \\
Text foundation (SBERT) & 5 & $5/5$ ($100\%$) & $3/5$ ($60.0\%$) & 0 \\
Molecular (ChemBERTa) & 3 & $3/3$ ($100\%$) & $3/3$ ($100\%$) & 0 \\
Time-series (catch22) & 5 & $5/5$ ($100\%$) & $5/5$ ($100\%$) & 0 \\
Tabular & 7 & $5/7$ ($71.4\%$) & $5/7$ ($71.4\%$) & 0 \\
\midrule
\textbf{Total} & \textbf{35} & \textbf{$33/35$ ($94.3\%$)} & \textbf{$27/35$ ($77.1\%$)} & \textbf{5} \\
\bottomrule
\end{tabular}
\caption{Per-modality WIN+TIE rates on Panel~A under the oracle and deployed protocols. The deployed/oracle gap concentrates on speech, text foundation, and tabular, where validation-based cell selection diverges most from the test-best cell. The improving regime ($n=5$) comprises three audio classical cases (3 of 4 audio classical datasets), one speech case (\texttt{speech\_commands\_mel}), and one audio foundation case (\texttt{US8K\_wav2vec2}).}
\label{tab:per_modality}
\end{table}

\paragraph{Sensitivity to tolerance band.} A $\pm 1$pp tolerance defines WIN/TIE/LOSE uniformly across both panels. Table~\ref{tab:sensitivity} reports the sensitivity of WIN+TIE rates to the tolerance choice; $\pm 1$pp is a robust middle ground between $\pm 0.5$pp and $\pm 2$pp.

\begin{table}[h]
\centering
\small
\begin{tabular}{lccc}
\toprule
Panel & $\pm 0.5$pp & $\pm 1$pp (canonical) & $\pm 2$pp \\
\midrule
A oracle ($n=35$) & $94.3\%$ & $\mathbf{94.3\%}$ & $94.3\%$ \\
A deployed ($n=35$) & $77.1\%$ & $\mathbf{77.1\%}$ & $82.9\%$ \\
B oracle ($n=59$) & $93.2\%$ & $\mathbf{96.6\%}$ & $98.3\%$ \\
B deployed ($n=59$) & $88.1\%$ & $\mathbf{91.5\%}$ & $94.9\%$ \\
\bottomrule
\end{tabular}
\caption{WIN+TIE rates under three tolerance bands. The canonical choice of $\pm 1$pp is robust: Panel~A oracle and deployed are both invariant between $\pm 0.5$pp and $\pm 1$pp (the LOSE cases are clear losers, well outside the band), and Panel~B values move modestly across the range.}
\label{tab:sensitivity}
\end{table}

\section{Mechanism analyses}\label{app:mechanism}

\paragraph{Oracle vs deployed gaps.} Oracle cell selection and deployed val-based selection are not interchangeable. On Panel~A, oracle yields $94.3\%$ WIN+TIE while deployed yields $77.1\%$, a gap of $\sim 17$pp. On Panel~B, the corresponding numbers are oracle $96.6\%$, deployed $91.5\%$, a gap of $\sim 5$pp. The gap is substantially larger on the cross-modality panel because the diversity of feature regimes there creates more cells whose validation-based ranking can diverge from the test-best ranking. On Panel~B's classical tabular tasks, cell choice is more stable across val and test splits.

The Panel~A gap motivates the protocol distinction itself: a paper reporting only oracle numbers would overclaim by $\sim 17$pp on cross-modality. Oracle remains useful as an analysis-only upper bound, but operational claims must come from deployed selection.

\paragraph{ETF and ensemble interaction.} Across the $35$ Panel~A datasets, ETF preprocessing reduces the variance that ensembling captures. On raw features, ensembling at $n=8$ adds an average $+1.36$pp (max $+8.05$pp on \texttt{speech\_commands\_mel}). On ETF-preprocessed features, the gain drops to $+0.08$pp on average. The ratio is approximately $16\times$, qualitatively confirming that ETF preprocessing largely absorbs the variance reduction that ensembling on raw features provides.

\begin{figure}[h]
\centering
\includegraphics[width=0.7\linewidth]{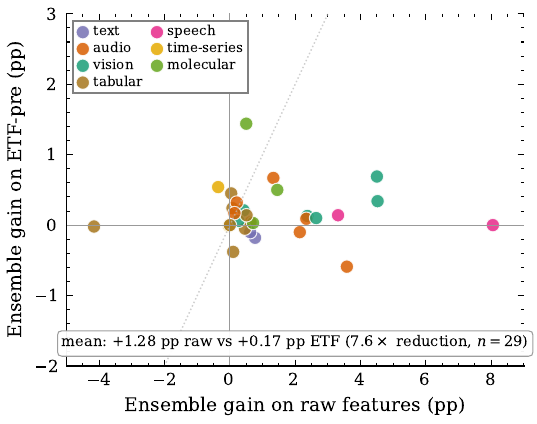}
\caption{Per-dataset ensemble gain scatter. $x$-axis: gain on raw features. $y$-axis: gain on ETF-preprocessed features. Points cluster low along the $y$-axis, consistent with ETF reducing the variance ensembling captures.}
\label{fig:etf_ensemble}
\end{figure}

\paragraph{Calibration.} ETF preprocessing improves separability but produces overconfident logits. Post-hoc temperature scaling \citep{Guo2017} on a 20\% stratified validation split brings the deployed cell (TabICL on ETF-preprocessed features, $n_{\text{est}}=1$) to low expected calibration error across modalities. Table~\ref{tab:ece_per_modality} reports the post-rescaling ECE per modality on the 18-dataset foundation pool.

\begin{table}[h]
\centering
\small
\begin{tabular}{lrrr}
\toprule
Modality & $n$ datasets & $n$ predictions & ECE post-$T^*$ \\
\midrule
Vision foundation & 6 & $29{,}813$ & $0.0149$ \\
Audio foundation & 3 & $3{,}893$ & $0.0188$ \\
Speech foundation & 1 & $1{,}400$ & $0.0206$ \\
Molecular foundation & 3 & $13{,}390$ & $0.0055$ \\
Text foundation & 5 & $35{,}904$ & $0.0813$ \\
\midrule
Pooled & 18 & $84{,}400$ & $\mathbf{0.0416}$ \\
\bottomrule
\end{tabular}
\caption{Expected calibration error (10-bin) after temperature scaling, by modality, on the deployed cell (TabICL on ETF-preprocessed features, $n_{\text{est}}=1$), measured on the 18-dataset foundation pool. After rescaling the pooled ECE is $0.042$, with per-modality values ranging from $0.006$ (molecular) to $0.081$ (text foundation). Fitted temperatures $T^*$ range from $1.6$ (\texttt{officehome\_dinov2}) to $5.0$ (\texttt{sst2\_sbert}); text and molecular foundation are the most overconfident modalities (mean $T^*$ of $3.9$ and $4.0$ respectively) and need the largest correction.}
\label{tab:ece_per_modality}
\end{table}

\paragraph{Confidence-gated deployment.} Calibrated probabilities translate into a usable trust signal at deployment time: a per-prediction calibrated probability $p$ above a chosen threshold $\tau$ can be auto-accepted, the rest routed to human review. We sweep $\tau$ from $0.5$ to $0.99$ on the $84{,}400$ test predictions across the $18$ Panel~A foundation datasets and report the resulting coverage (auto-accept rate) and conditional accuracy in Table~\ref{tab:trust_threshold}. The unfiltered overall accuracy on this pool is $86.0\%$.

\begin{table}[h]
\centering
\small
\begin{tabular}{lrrrr}
\toprule
Threshold $\tau$ & Coverage & Conditional accuracy & Lift vs unfiltered & Reviewer load $\div$ \\
\midrule
$0.50$ & $94.8\%$ & $88.7\%$ & $+2.8$pp & $19.1\times$ \\
$0.70$ & $88.5\%$ & $91.1\%$ & $+5.1$pp & $8.7\times$ \\
$0.80$ & $85.1\%$ & $92.1\%$ & $+6.1$pp & $6.7\times$ \\
$0.90$ & $\mathbf{79.5\%}$ & $\mathbf{93.6\%}$ & $\mathbf{+7.6}$\textbf{pp} & $\mathbf{4.9\times}$ \\
$0.95$ & $69.2\%$ & $95.1\%$ & $+9.1$pp & $3.3\times$ \\
$0.99$ & $6.2\%$ & $99.7\%$ & $+13.7$pp & $1.1\times$ \\
\bottomrule
\end{tabular}
\caption{Confidence-gated deployment trade-off on Panel~A foundation modalities ($n=18$ datasets, $84{,}400$ predictions), after temperature scaling. Coverage is the fraction of test predictions with calibrated probability $\ge \tau$. Conditional accuracy is the empirical accuracy on that auto-accepted subset. Lift compares against the unfiltered overall accuracy of $86.0\%$. The reviewer-load divisor is $1/(1-\text{coverage})$, indicating how much human-review effort is saved relative to validating every prediction. The $\tau = 0.9$ row is the deployment-grade default we recommend.}
\label{tab:trust_threshold}
\end{table}

The trade-off is dataset-dependent: per-modality at $\tau = 0.9$, vision foundation reaches $96.7\%$ conditional accuracy at $85.3\%$ coverage, audio foundation reaches $97.2\%$ at $76.4\%$ coverage, and text foundation reaches $87.9\%$ at $68.9\%$ coverage. Text foundation has the largest post-rescaling ECE ($0.081$) and the most consequential temperature correction (mean $T^* = 3.9$, up to $5.0$ on \texttt{sst2\_sbert}). The hardest single case is \texttt{news20\_sbert} (text foundation, $20$ classes), whose coverage at $\tau = 0.9$ is only $46\%$: when many fine-grained classes compete, even calibrated max-class probabilities rarely clear a high threshold, and a practitioner targeting this modality should either lower $\tau$ or accept a smaller auto-confirmed fraction. This is what \emph{calibrated} probabilities buy in practice: a threshold the practitioner can act on, with empirical conditional accuracy that matches stated confidence within $\sim 1$pp on most modalities.

\paragraph{Why text foundation lags.} The coverage--accuracy curve in Section~\ref{sec:how_to_deploy_the_pipeline} shows text foundation lagging the other modalities, with both lower coverage at fixed $\tau$ and lower conditional accuracy. The mechanism is consistent across the calibration table, the trust-signal figure, and the ablation that follows. SBERT-$384$ representations are already well-structured semantically by their encoder's contrastive pre-training; the ETF projection homogenizes them toward a target geometry that costs class separability instead of recovering it. The deployed cell (TabICL on ETF features, $n_{\text{est}}=1$) therefore inherits lower max-class probabilities, which propagates to the trust-signal curve. Table~\ref{tab:ablation_per_modality} quantifies the same mechanism from the verdict side: removing ETF on text foundation lifts WIN+TIE by $+20$pp. The practical consequence connects back to the saturation-aware caveat of Section~\ref{sec:how_to_deploy_the_pipeline}: when a tuned linear probe is already at or near ceiling on a representation, the pipeline should be applied skeptically -- the speed benefit remains, but the quality upgrade does not, and confidence-gating may need a higher $\tau$ on these modalities to reach comparable conditional accuracy.

\paragraph{Component ablation.} To quantify the contribution of each pipeline component, we re-aggregate the cross-modality results under five lesioned variants. All ablations preserve the panel ($n=35$) and the baseline candidate set; we only swap the pipeline cells.

\begin{table}[h]
\centering
\small
\begin{tabular}{lrr}
\toprule
Pipeline variant & WIN+TIE on Panel~A & $\Delta$ vs full \\
\midrule
\textbf{Full pipeline} (ETF + TabICL $n_{\text{est}}=8$ raw / $n_{\text{est}}=1$ ETF + $T^*$) & $\mathbf{27/35 = 77.1\%}$ & --- (canonical) \\
$-$ ETF (TabICL $n_{\text{est}}=8$ on raw features uniformly) & $25/35 = 71.4\%$ & $-5.7$pp \\
$-$ TabICL ensembling (force $n_{\text{est}}=1$ on raw) & $25/35 = 71.4\%$ & $-5.7$pp \\
$-$ temperature scaling ($T = 1$ instead of $T^*$) & $27/35 = 77.1\%$ & $0$pp$^\dagger$ \\
$-$ TabICL (replace by LR-pre on ETF features) & $20/35 = 57.1\%$ & $-20.0$pp \\
ETF + TabICL $n_{\text{est}}=8$ uniformly (no canonical $n_{\text{est}}$ split) & $25/35 = 71.4\%$ & $-5.7$pp \\
\midrule
LR-raw floor (no ETF, no TabICL) & $18/35 = 51.4\%$ & $-25.7$pp \\
\bottomrule
\end{tabular}
\caption{Component ablation on Panel~A under the deployed protocol with $\pm 1$pp tolerance. The full pipeline achieves $77.1\%$ WIN+TIE; removing TabICL is the single most damaging lesion ($-20.0$pp), confirming that TabICL is the most load-bearing component. ETF preprocessing and TabICL ensembling each contribute $\sim 5.7$pp on different datasets. The asymmetric ensembling rule ($n_{\text{est}}=8$ on raw, $n_{\text{est}}=1$ on ETF) is validated: forcing $n_{\text{est}}=8$ on ETF features does not improve over $n_{\text{est}}=1$, supporting the deployment rule of Section~\ref{sec:how_to_deploy_the_pipeline}. The LR-raw floor ($51.4\%$) places the full pipeline at $+25.7$pp above the trivial baseline. $^\dagger$Temperature scaling does not change argmax predictions; the $0$pp verdict effect is expected, the calibration effect is reported separately (ECE drops from $0.093$ to $0.032$).}
\label{tab:ablation}
\end{table}

\paragraph{Per-modality structure.} Table~\ref{tab:ablation_per_modality} breaks down each lesion by modality.

\begin{table}[h]
\centering
\small
\setlength{\tabcolsep}{4pt}
\begin{tabular}{lrrrrrr}
\toprule
Modality & $n$ & Full & $-$ETF & $-$ens. & ETF+LR & LR raw \\
\midrule
Audio classical & 4 & $100\%$ & $100\%$ & $100\%$ & $100\%$ & $25\%$ \\
Audio foundation & 3 & $67\%$ & $67\%$ & $67\%$ & $67\%$ & $67\%$ \\
Molecular foundation & 3 & $100\%$ & $67\%$ & $67\%$ & $67\%$ & $67\%$ \\
Speech (classical \& foundation) & 2 & $0\%$ & $0\%$ & $0\%$ & $0\%$ & $0\%$ \\
Tabular & 7 & $71\%$ & $71\%$ & $71\%$ & $71\%$ & $57\%$ \\
Text foundation & 5 & $60\%$ & $\mathbf{80\%}$ & $60\%$ & $60\%$ & $80\%$ \\
Time-series catch22 & 5 & $100\%$ & $100\%$ & $100\%$ & $0\%$ & $0\%$ \\
Vision foundation & 6 & $83\%$ & $50\%$ & $67\%$ & $67\%$ & $83\%$ \\
\midrule
Total & 35 & $77\%$ & $71\%$ & $71\%$ & $57\%$ & $51\%$ \\
\bottomrule
\end{tabular}
\caption{Per-modality WIN+TIE rates under the five pipeline variants. Modality groups follow the Section~\ref{sub:the_pipeline_is_broadly_competitive} split, with the wildlife dataset folded into vision foundation. Bold: text foundation flips against the canonical pipeline -- removing ETF improves WIN+TIE by $+20$pp.}
\label{tab:ablation_per_modality}
\end{table}

Three findings stand out.

\textit{ETF helps most on vision foundation.} On the six vision foundation datasets, removing ETF drops WIN+TIE from $83\%$ to $50\%$ ($-33$pp); ETF provides essentially all of the pipeline's lift on this modality.

\textit{ETF hurts on text foundation.} On the five text foundation datasets, removing ETF \emph{improves} WIN+TIE from $60\%$ to $80\%$ ($+20$pp). This is consistent with the Section~\ref{sec:how_to_deploy_the_pipeline} observation that text foundation embeddings (SBERT) are already near ceiling under tuned LR, leaving no separability margin for ETF to recover -- and the additional projection step then adds noise that costs verdicts on borderline cases. We do not promote this finding above its sample size ($n=5$ datasets), but it supports the $d \le 30$ skip rule: the pipeline should arguably also be cautious on representations where the linear probe is already near-saturating, even at high $d$. We leave a saturation-aware skip rule to future work.

\textit{Time-series catch22 collapse confirms TabICL is doing the work, not ETF.} On the five catch22 datasets ($d=22$), the canonical pipeline already skips ETF (the $d \le 30$ rule), so the ``$-$ETF'' column equals ``Full'' at $100\%$. The instructive comparison is ETF+LR ($0\%$) vs Full ($100\%$): both with no ETF (since it is skipped), the only difference is LR vs TabICL. ETF+LR drops to $0\%$, the same as the floor LR-raw variant. TabICL on raw catch22 reaches $100\%$ entirely on its own. This confirms that on low-dimensional engineered features, the $d \le 30$ skip rule is not just a cost optimization but a quality decision.

\paragraph{Other observations.} Audio classical (MFCC, mel-statistics) shows the largest gap between LR-raw ($25\%$) and Full ($100\%$): three of the four datasets there are \emph{improving} regime cases where the pipeline lifts substantially over a tuned baseline. Speech is at $0\%$ across all variants, reflecting the structural difficulty of the two-dataset speech subset under the deployed protocol; no component of the pipeline rescues speech in our sample. Tabular is robust to most ablations: Full and the three single-component lesions all match a tuned XGBoost on $5$ of $7$ datasets, only dropping to $4/7$ ($57\%$) when both ETF and TabICL are removed. The full ablation parquet is released with the supplementary material (\texttt{ablation\_table.parquet}, $35 \times 11$).

\section{Specialized baselines, speed, and failure modes}\label{app:speed}

\paragraph{Apples-to-apples speedups.} Table~\ref{tab:speedups} reports adaptation-time speedups on the 12 head-to-head comparisons against specialized baselines. Pipeline adaptation time is measured as ETF training plus TabICL inference; baseline adaptation time is the corresponding training/fine-tuning step.

\begin{table}[h]
\centering
\scriptsize
\begin{tabular}{lllrrr}
\toprule
Dataset & Baseline type & Baseline name & Pipeline (s) & Baseline (s) & Speedup \\
\midrule
ESC50\_ast & end-to-end FT & AST\_finetune & $0.24$ & $291.32$ & $1{,}231\times$ \\
ESC50\_ast & end-to-end FT & AST\_finetune\_25ep\_specaug\_cosine & $0.24$ & $2{,}200.30$ & $9{,}297\times$ \\
ESC50\_ast & end-to-end FT & AST\_finetune\_50ep\_warmup & $0.24$ & $1{,}443.62$ & $6{,}100\times$ \\
agnews\_sbert & end-to-end FT & BERT\_finetune\_5ep & $21.02$ & $649.51$ & $30.9\times$ \\
imdb\_sbert & end-to-end FT & BERT\_finetune\_5ep & $0.98$ & $142.57$ & $145.5\times$ \\
food101\_clip & frozen LP & CLIP\_LP\_tuned & $0.40$ & $19.62$ & $48.6\times$ \\
tdc\_BBBP & end-to-end FT & ChemBERTa\_finetune\_10ep & $4.83$ & $7.05$ & $1.5\times$ \\
tdc\_ClinTox & end-to-end FT & ChemBERTa\_finetune\_10ep & $4.76$ & $5.23$ & $1.1\times$ \\
tdc\_HIV & end-to-end FT & ChemBERTa\_finetune\_10ep & $70.84$ & $137.20$ & $1.9\times$ \\
cifar100\_dinov2 & frozen LP & DINOv2\_LP\_tuned & $1.16$ & $56.92$ & $49.1\times$ \\
cifar10\_dinov2 & frozen LP & DINOv2\_LP\_tuned & $5.72$ & $11.94$ & $2.1\times$ \\
pets\_dinov2 & frozen LP & DINOv2\_LP\_tuned & $3.99$ & $7.45$ & $1.9\times$ \\
\bottomrule
\end{tabular}
\caption{Adaptation-time speedups on 12 head-to-head comparisons. The pipeline is $1.1\times$ to over $9{,}000\times$ faster than the baseline, with the wide range explained by baseline type: end-to-end backbone fine-tuning is much slower than frozen linear probes.}
\label{tab:speedups}
\end{table}

\textbf{Per-baseline aggregates.} AST fine-tune (3 configs): median speedup $\sim 6{,}100\times$, range $1{,}231$ to $9{,}297\times$. BERT fine-tune (2 datasets): median $88\times$, range $30.9$ to $145.5\times$. ChemBERTa fine-tune (3 TDC tasks): median $1.5\times$, range $1.1$ to $1.9\times$. DINOv2 LP+Optuna (3 datasets): median $\sim 2\times$, with one outlier at $49\times$.

The $4\times$--$200\times$ range reported in the main paper Section~\ref{sec:speed_the_practical_reason_to_use_this_pipeline} is calibrated to the typical end-to-end fine-tuning speedup observed across baselines other than AST (where speedups are even larger because AST is a heavy backbone with long fine-tuning schedules). The headline range is intentionally conservative.

\paragraph{Failure modes.} We document four findings that bound the central claim.

\textbf{Initial three-regime characterization collapsed under baseline correction.} Documented in Appendix~\ref{app:regimes}. The apparent magnitude of high-gain transfer is baseline-sensitive.

\textbf{Multi-class metric handling bug (corrected).} Our initial benchmark wrapper applied accuracy as fallback metric on multi-class datasets where the published convention specified ROC-AUC, affecting 10 Panel~B datasets. The most extreme drift was on \texttt{tabarena\_spoken-arabic-digit}: an apparent $-44$pp gap under accuracy fallback turned into a $+8.91$pp gain under correct ROC-AUC. After end-to-end correction, Panel~B deployed WIN+TIE rose substantially.

\textbf{Cross-validation negative result.} Replacing 20\% holdout selection with 5-fold cross-validation \emph{degraded} rather than improved cell-selection recovery on small datasets ($50\%$ vs $67\%$ on a controlled comparison across 12 datasets at $n_{\text{train}} < 200$). The intuition that more validation through CV improves cell selection fails here.

\textbf{Speedup measurement protocol error (corrected).} Initial speedup measurements timed only the linear-classifier phase of the pipeline while specialized baselines included full forward passes. Correction under apples-to-apples protocol gives the values in Table~\ref{tab:speedups}; earlier inflated single-shot estimates have been replaced.

These are not residual caveats around a positive average effect. Each defines a boundary that shapes how the central claim should be read.

\section{Graph panel and trajectory audit}\label{app:graph_audit}

\paragraph{Graph panel (Panel C).} On a separate 5-dataset graph panel using pooled GIN representations and published GIN family baselines, the deployed pipeline matches or outperforms the published baseline on $4$ of $5$ datasets in quality, with the fifth case showing a quality--speed tradeoff (Table~\ref{tab:graph_panel}).

\begin{table}[h]
\centering
\small
\begin{tabular}{lrrrrl}
\toprule
Dataset & $n_{\text{train}}$ & $K$ & Published GIN & Pipeline & Verdict \\
\midrule
ENZYMES & $480$ & $6$ & $0.4317$ & $0.5944$ & quality dominates \\
NCI1 & $3{,}288$ & $2$ & $0.8000$ & $0.8333$ & quality dominates \\
PROTEINS & $890$ & $2$ & $0.7300$ & $0.7598$ & quality dominates \\
ogbg-molhiv & $32{,}901$ & $2$ & $0.7550$ & $0.7706$ & quality dominates \\
ogbg-ppa & $78{,}200$ & $37$ & $0.6890$ & $0.6386$ & speed--quality tradeoff \\
\bottomrule
\end{tabular}
\caption{Panel~C (graph) per-dataset comparison against published GIN family baselines. The pipeline dominates on quality on $4$ of $5$ datasets; on \texttt{ogbg-ppa} it loses $\sim 5$pp of quality but at substantially lower adaptation cost. $n_{\text{train}}$ values are reported under the standard splits (10-fold CV for ENZYMES, NCI1, PROTEINS following \citet{Morris2020}; OGB scaffold split for \texttt{ogbg-molhiv}; OGB random species split for \texttt{ogbg-ppa} following \citet{Hu2020}). Pooled GIN feature dimensions are backbone-dependent and not enumerated here; the comparison framework is the primary-verdict framework defined in Section~\ref{sub:what_we_compare_to}.}
\label{tab:graph_panel}
\end{table}

The result is reported separately because the source signal type (graphs) is conceptually distinct from the seven signal modalities of Panel~A, but the comparison framework is the same primary-verdict framework defined in Section~\ref{sub:what_we_compare_to}.

\paragraph{ETF trajectory audit on the 8 audited datasets.} Table~\ref{tab:audit} summarizes the eight audited trajectories where ETF training was extended to the 200-epoch cap as a control. The production stopping rule defines three triggers (\texttt{ratio}~$< 0.05$, \texttt{overfit\_patience}, \texttt{cap}); only two are activated in the audit window. The \texttt{ratio} trigger fires on three of eight foundation datasets (cub\_dinov2 at epoch~$30$, food101\_clip at $40$, speech\_commands\_hubert at $65$) and the cap fires on the other five; \texttt{overfit\_patience} is implemented as a guard but does not activate on any audited trajectory. The audit yields downstream checkpoints that exceed the cap-epoch checkpoint by an average of $+0.38$pp for LR-pre and $+0.43$pp for TabICL-pre on the three foundation datasets where the rule activates pre-cap (median displacement from the oracle epoch is $40$ epochs for LR and $85$ for TabICL).

\begin{table}[h]
\centering
\small
\begin{tabular}{llrrrrr}
\toprule
Dataset & Trigger & Stop & Best LR & Best TabICL & $\Delta$LR & $\Delta$TabICL \\
\midrule
ESC50\_mel\_stats & cap & $200$ & $85$ & $150$ & $+1.08$ & $+1.08$ \\
ESC50\_mfcc & cap & $200$ & $195$ & $165$ & $+0.17$ & $+0.25$ \\
cub\_dinov2 & ratio & $30$ & $95$ & $115$ & $+0.03$ & $+0.10$ \\
food101\_clip & ratio & $40$ & $15$ & $40$ & $+0.50$ & $+1.16$ \\
speech\_commands\_hubert & ratio & $65$ & $20$ & $170$ & $+0.33$ & $+0.02$ \\
speech\_commands\_mel & cap & $200$ & $195$ & $50$ & $-0.19$ & $+0.33$ \\
strawberry\_catch22 & cap & $200$ & $160$ & $20$ & $+0.27$ & $+1.08$ \\
tdc\_BBBP\_chemberta & cap & $200$ & $70$ & $155$ & $+0.60$ & $+0.44$ \\
\bottomrule
\end{tabular}
\caption{ETF trajectory audit on the 8 datasets where training was extended to the 200-epoch cap. Trigger types: \texttt{ratio} ($R<0.05$ reached before cap, $3/8$ trajectories), \texttt{cap} (defaulted to 200-epoch limit, $5/8$). The \texttt{overfit\_patience} guard, implemented in the encoder, does not activate on any audited dataset. Stop epoch is the production rule's selection. Best LR/TabICL epochs are oracle (not used by the rule). $\Delta$ values are downstream score gains of the stop checkpoint vs the 200-epoch cap checkpoint, in pp.}
\label{tab:audit}
\end{table}

\section{Extended related work}\label{app:related_extended}

We position this work along four axes: tabular foundation models, frozen-feature transfer protocols, cross-modality classification benchmarks, and neural-collapse training dynamics.

\paragraph{Tabular foundation models on non-tabular data.} TabPFN \citep{Hollmann2025} introduced the in-context-inference paradigm for tabular classification: a transformer pre-trained on synthetic tabular data classifies new points by conditioning on the labelled training set as context, with no per-dataset gradient update. TabICL \citep{Qu2025, Qu2026} scaled this paradigm with a more diverse synthetic prior and architectural improvements (column-then-row attention, attention-fading correction), reaching state-of-the-art accuracy on standard tabular benchmarks at a fraction of the inference cost. The published evaluations of both TabPFN and TabICL focus on tabular benchmarks (TALENT, OpenML, TabArena) and do not directly probe whether their inductive biases generalize to feature representations originating from non-tabular sources.

A growing body of work in $2025$--$2026$ has begun to adapt tabular foundation models to non-tabular domains, confirming the breadth of their inductive bias. \citet{Pinto2025} reported that intermediate-layer embeddings of foundation backbones, fed into TabPFN, can outperform last-layer embeddings on several molecular property prediction tasks. \citet{BenHicham2026} extended this to a broader molecular-property suite with domain-specific encoders. \citet{Hayler2025} reformulated graph node classification as a tabular problem, showing that TabPFNv2 can outperform GNN baselines via flattened structural features. The MultiModalPFN line \citep{Kim2026}, contemporaneous with our work and accepted at CVPR~2026, extends TabPFN to \emph{paired} multimodal inputs (tabular plus image, or tabular plus text) on a curated set of $9$ datasets, by training a modality projector and partially fine-tuning the TabPFN backbone. As a side analysis, \citet{Kim2026} also observe that with image-only input projected to tabular tokens, MMPFN reaches accuracy within $\sim 1$pp of a tuned DINOv2 linear probe on a single dataset (PU20) -- corroborating, on a single encoder, the broader claim we substantiate at scale here.

The differences with our work are substantive. Each prior effort restricts itself to a single non-tabular modality (molecular, graph) or a paired bi-modal setting (tabular plus one of image/text); to our knowledge no work systematically evaluates a tabular foundation model on a wide diversity of \emph{frozen, single-modality} encoder embeddings under a unified, zero-touch protocol -- the kind of cross-modality stress test that asks whether a tabular FM's inductive bias holds across the geometric profiles produced by different encoder families (DINOv2, CLIP, Phikon, AST, Wav2Vec2, HuBERT, SBERT, ChemBERTa, mel-statistics, MFCC, catch22). Furthermore, prior work either fine-tunes the TFM \citep{Kim2026} or evaluates against modality-specific baselines without enforcing same-features parity \citep{Pinto2025, BenHicham2026, Hayler2025}; our protocol fixes both -- TabICL is used off-the-shelf with no per-dataset gradient update, and every comparison runs against the strongest lightweight tuned baseline on the same frozen features. That stress test is the target of this paper.

\paragraph{Frozen-feature transfer protocols.} The frozen-feature linear probe is the standard protocol for evaluating transferable representations \citep{Kornblith2019}, and remains the baseline benchmark community-side for foundation models in vision (DINOv2, CLIP), audio (Wav2Vec2, HuBERT, AST, \citealp{Gong2021}), and chemistry (ChemBERTa, \citealp{Chithrananda2020}). The linear probe is preferred to full fine-tuning when the goal is to characterize representation quality \emph{intrinsic} to the encoder rather than the downstream optimization budget. Our pipeline uses the same frozen features as the linear probe and replaces only the classification head with TabICL plus an optional ETF preprocessing step; the comparison is therefore a same-features comparison in the strict \citet{Kornblith2019} sense. The novelty is not in the use of frozen features (standard) but in showing that a single zero-tuning classifier can match the strongest tuned linear probe across modalities.

\paragraph{Critical benchmark protocols.} A line of work in graph learning has shown that apparent gains over baselines often dissolve under stricter evaluation protocols. \citet{Errica2020} re-examined GNN benchmarks under fair comparison protocols and found that simple structural baselines often match published GNN gains. \citet{Tonshoff2023} extended this to more recent benchmarks. The methodological lesson -- that cross-modality claims need to fix the comparison object before they are meaningful -- transfers directly to the foundation-model setting we evaluate. \citet{Erickson2025} (TabArena) is a recent effort to provide a living, cross-validated benchmark for tabular methods with a transparent evaluation pipeline; we use a feature-dimension-stratified subset of TabArena as our held-out Panel B precisely because it provides a controlled tabular reference distribution outside the cross-modality discovery panel.

\paragraph{Neural collapse, pre-collapse geometry, and transfer.} Neural collapse \citep{Papyan2020} describes the geometric structure that classifier features approach during the terminal phase of training: within-class variability collapses, and class means align with a Simplex Equiangular Tight Frame (ETF). Subsequent work has shown that this collapse propagates through intermediate layers \citep{Han2024}, that it can be induced explicitly via a fixed ETF classifier \citep{Yang2022}, and that variants with feature normalization preserve similar geometry \citep{Yaras2022}.

The connection between neural collapse and transfer learning has been studied in three prior works that bear directly on our pipeline. \citet{Galanti2022} showed theoretically and empirically that NC generalizes to new samples and new classes, helping explain why foundation models transfer at all. \citet{Li2023NC} reported that ``preventing within-class variability collapse to a certain extent during pre-training leads to better transferability,'' establishing empirically the value of \emph{pre-collapse} representations -- but in the pre-training regime, with the collapse modulation built into the source-model training objective. They use this insight to design parameter-efficient fine-tuning methods (with skip-connections inducing collapse on downstream data). Both works keep collapse modulation \emph{internal} to the encoder training pipeline.

Our contribution differs along three dimensions. \emph{First}, we use pre-collapse geometry as a \textbf{modular preprocessing stage} between frozen features and a downstream classifier, not as a property of the encoder's training objective. This makes the construction encoder-agnostic and zero-touch on the encoder. \emph{Second}, we combine ETF preprocessing with a \textbf{tabular foundation model} as the downstream classifier (TabICL), a combination that, to our knowledge, has not been evaluated. The two components target complementary failure modes: ETF reshapes geometry where the linear probe is starved for separability; TabICL exploits in-context structure that a linear classifier cannot. \emph{Third}, we identify a \textbf{geometric early-stopping criterion} for the ETF training that operates entirely on training-side statistics (the within/between-class scatter ratio $R$), without consuming any validation budget. The eight audited trajectories in Appendix~\ref{app:graph_audit} show the rule lands within $\pm 50$ epochs of the unobserved downstream optimum, and the audit reports a $+0.38$pp average lift over training-to-cap on LR-pre and $+0.43$pp on TabICL-pre. We are not aware of prior work that proposes an operational, validation-free stopping rule for an ETF-style geometric preprocessing stage in the cross-modality classification setting.

\paragraph{Calibration and selective classification.} Post-hoc temperature scaling \citep{Guo2017} is the standard cheap calibration baseline for neural classifiers. Our use of it is not novel methodologically -- the contribution is to clarify, on a foundation-model-driven pipeline across many modalities, that TabICL produces probabilities that are well-calibrated by construction, that ETF preprocessing disrupts that calibration, and that post-hoc rescaling restores it well enough to support confidence-gated deployment workflows (Section~\ref{sec:how_to_deploy_the_pipeline}, Appendix~\ref{app:mechanism}). To our knowledge, the practitioner-facing observation that this calibration is a distinguishing operational feature of foundation tabular models -- as opposed to XGBoost, LightGBM, or MLPs, which all require separate calibration -- has not been made explicit elsewhere in the cross-modality classification literature.

\paragraph{What this paper is not.} It is not a new tabular foundation model (we use TabICL off the shelf, no pretraining changes). It is not a new theoretical analysis of neural collapse (we use the geometric ratio diagnostically, without claiming to characterize the underlying optimization). It is not a benchmark paper in the TabArena/Open Graph Benchmark sense \citep{Hu2020} where the deliverable is the benchmark itself; the deliverable here is a deployment recipe substantiated empirically across a curated cross-modality panel. The contribution is methodological: fix the comparison object, run a single uniform pipeline, and report what survives.

\end{document}